\documentclass[12pt]{article}
\usepackage{hyperref}
\usepackage{graphicx}
\usepackage{float}
\usepackage{setspace}
\usepackage{lineno}
\usepackage{amsmath}
\usepackage{amssymb}
\usepackage{subcaption}
\usepackage[numbers, square, sort&compress]{natbib}
\usepackage{authblk}
\usepackage{geometry}
\usepackage{tabularx}
\usepackage{booktabs}
\usepackage{multirow}
\usepackage{array}
\usepackage{makecell}
\usepackage{longtable}
\usepackage{pdflscape}
\usepackage{afterpage}
\usepackage{rotating}
\usepackage{lscape}
\usepackage{tabularx}
\usepackage{booktabs}
\usepackage{listings}
\usepackage{makecell}
\usepackage{tcolorbox}
\usepackage{enumitem}

\usepackage{xcolor}

\begin{document}

\title{Thematic Analysis with Open-Source Generative AI and Machine Learning: A New Method for Inductive Qualitative Codebook Development}

\author[1]{Andrew Katz}

\author[1]{Gabriella Coloyan Fleming}

\author[2]{Joyce Main}

\affil[1]{Department of Engineering Education, Virginia Tech}
\affil[2]{School of Engineering Education, Purdue University}

\date{}

\maketitle

\begin{abstract}
This paper aims to answer one central question: \textit{to what extent can open-source generative text models be used in a workflow to approximate thematic analysis in social science research?} To answer this question, we present the Generative AI-enabled Theme Organization and Structuring (GATOS) workflow, which uses open-source machine learning techniques, natural language processing tools, and generative text models to facilitate thematic analysis. To establish validity of the method, we present three case studies applying the GATOS workflow, leveraging these models and techniques to inductively create codebooks similar to traditional procedures using thematic analysis. Specifically, we investigate the extent to which a workflow comprising open-source models and tools can inductively produce codebooks that approach the known space of themes and sub-themes. The problem motivating the workflow is simple: in many social science research settings, researchers and stakeholders generate large amounts of text. Key insights into phenomena might reside in that collection of text, but the volume of data is too large to analyze by hand. Examples of such data include hundreds of hours of audio recordings, thousands of written documents, and responses to open-ended questions on surveys. To address the challenge of gleaning insights from these texts, we combine open-source generative text models, retrieval-augmented generation, and prompt engineering to identify codes and themes in large volumes of text, i.e., generate a qualitative codebook. The process mimics an inductive coding process that researchers might use in traditional thematic analysis by reading text one unit of analysis at a time, considering existing codes already in the codebook, and then deciding whether or not to generate a new code based on whether the extant codebook provides adequate thematic coverage. We demonstrate this workflow using three synthetic datasets from hypothetical organizational research settings: a study of teammate feedback in teamwork settings, a study of organizational cultures of ethical behavior, and a study of employee perspectives about returning to their offices after the pandemic. We show that the GATOS workflow is able to identify themes in the text that were used to generate the original synthetic datasets. We conclude with a discussion of the implications of this work for social science research and the potential for open-source generative text models to facilitate qualitative data analysis.

\end{abstract}


\section{Introduction}

In social science research, one may want to answer questions about the experiences, beliefs, and attitudes of individuals. To do so, the researcher can collect quantitative and/or qualitative data. Qualitative and quantitative methodologies have their own underlying epistemological assumptions and traditions \cite{salomon1991transcending,halfpenny1979analysis}. They also come with tradeoffs. While there are many mathematically derived ways to analyze quantitative data, qualitative data has a different foundation and set of approaches for analysis. One form of qualitative data analysis involves a process called coding. Coding is the process of identifying themes, patterns, and insights in a corpus of text \cite{saldana2011fundamentals}. This process is often done manually, but it can be time-consuming and labor-intensive. Moreover, it does not scale well to large volumes of text. Whereas a small group of researchers can analyze tens or even hundreds of observations, thousands or tens of thousands present unique challenges for resources and consistency. These issues facing qualitative researchers in social research who want to work on a large scale are the central challenges motivating this paper. In our work, we aim to validate a method that uses modern machine learning techniques, especially open-source large language models, to analyze large volumes of text and ultimately generate qualitative codebooks.

Our approach for doing this validation work is simple: compare our new method and its resulting codebooks with the themes and sub-themes that were intentionally built into synthetic data generated for this study. To do this, we simulate three datasets that imitate common data collection and contexts in organizational research. The first dataset is designed to mimic a study of teamwork dynamics, specifically teammate feedback, which is a common research topic in organizational behavior and management research \cite{wu2014feedback,potosky2022modeling,dominick1997effects,donia2015peer}. The second dataset is motivated by studies of organizational cultures of ethical behavior \cite{key1999organizational,roy2024ethical,debode2013assessing,kaptein2011understanding,kuye2013organizational}. The third dataset is inspired by more recent research on employee perspectives about returning to their offices after the COVID-19 pandemic \cite{gibson2023should,wang2021returning,liu2020don,hou2021study}. In the following sections, we will show that our new method that leverages open-source models and tools is able to identify themes in these qualitative datasets reliably and efficiently. The ability to perform this kind of text analysis for inductively generative qualitative codebooks is a significant contribution to the organizational research literature and social science research more broadly. Before describing the method, we briefly review literature on qualitative coding and large language models in social science research.

\section{Background}

\subsection{Qualitative Codebooks and Coding}
Qualitative coding has a long tradition in social science research. It is a process of identifying themes, patterns, and insights in a corpus of text \cite{saldana2011fundamentals}. Thematic analysis is one of the foundational methods in qualitative data analysis \cite{braun2006using}. Benefits of thematic analysis are that it can be flexibly utilized in a broad range of approaches (e.g., theoretical, conceptual, epistemological) while providing detailed, complex findings. Following the Braun and Clarke tradition, thematic analysis has six phases: ``1) familiarizing yourself with the data, 2) generating initial codes, 3) searching for themes, 4) reviewing themes, 5) defining and naming themes, and 6) producing the report'' \citep[page 87]{braun2006using}. Codebooks begin to be developed in phase two, refined in phases three through five, and used in phase six. A final codebook shows all of the codes for an analyzed data set organized hierarchically to show the relationships between codes (e.g., themes and sub-themes) and descriptions and examples for each code \cite{gery2000data}. Of course, this is just one example of qualitative coding. Many more are outlined in popular qualitative coding manuals \cite{saldana2021coding}. Regardless of the tradition, there are non-trivial questions about what constitutes a code, differences between codes and themes, and which procedure is most advantageous for a given research question \cite{elliott2018thinking}.

The coding process is often done manually, but it can be time-consuming and labor-intensive. Moreover, it does not scale well to large volumes of text; whereas a small group of researchers can analyze tens or even hundreds of units of text, thousands or tens of thousands present unique challenges for resources and consistency. These issues facing qualitative researchers working on a large scale are the central challenges motivating this paper. In this work, we aim to validate a method that uses modern machine learning techniques to code large volumes of text.

\subsection{Natural Language Processing and Machine Learning in Thematic Analysis}

There have been several recent attempts over the past 15 years to accomplish this task of identifying themes in large corpora of texts. These attempts have used a variety of machine learning techniques, including topic modeling, clustering, and deep learning. For example, \cite{blei2003latent} used a technique called latent Dirichlet allocation (LDA) to identify topics in a corpus of text. A more modern version of LDA that uses transformer-based language models is BERTopic \cite{grootendorst2022bertopic}. Before these deep neural network approaches, however, computer-assisted qualitative data analysis software was around for decades \cite{cope2014computer,kelle1995computer}. These tools, such as NVivo, ATLAS.ti, and Dedoose, have been used to facilitate the coding process by providing a digital interface for researchers to organize and analyze their data. Other tools like quanteda \cite{benoit2018quanteda} or tidytext \cite{silge2016tidytext} in R have been used to analyze text data in a more programmatic way, but these are not designed for thematic analysis and instead focus more on text mining. Until the past few years, there were glimmers of potential for using machine learning techniques to facilitate thematic analysis, but the computational resources and models were not yet at a level of sophistication to approximate anything close to human-level coding.

Since the release of ChatGPT in November 2022, large language models (LLMs) have pervaded global discourse. Unsurprisingly, qualitative researchers immediately began exploring the use of LLMs for qualitative data analysis (QDA). ChatGPT is one of the most commonly used LLMs for QDA and has been used to study a variety of topics, including universal basic income, PhD students' transitions to being independent researchers, and engineering student career interests \cite{hamilton2023exploring,tai2024examination,katz2023utility,Katz2024using}. 

While some papers focus on the application of LLMs to specific use cases (i.e., data sets), others undertake establishing workflows for utilizing LLMs for QDA \cite{de2024performing,gao2024collabcoder,katz2023utility}. As this paper's methods employ thematic analysis, we center our literature review on that particular method of QDA. De Paoli (2024) and Katz et al. (2023) focused on human-in-the-loop LLM-assisted thematic analysis with an individual human colder, while Gao et al. (2024) developed an LLM-assisted collaborative coding platform for multiple researchers. Table \ref{tab:llm-assisted-thematic-analysis} shows a comparison of these three workflows and the approaches they are grounded in.

\begin{table}[htbp]
    \centering
    \fontsize{10}{12}\selectfont
    \caption{Comparison of LLM-assisted thematic analysis workflows}
    \begin{tabularx}{\textwidth}{>{\raggedright\arraybackslash}p{0.3\textwidth} >{\raggedright\arraybackslash}p{0.3\textwidth} >{\raggedright\arraybackslash}X}
    \toprule
    \textbf{De Paoli (2024)} & \textbf{Katz et al. (2023)} & \textbf{Gao et al. (2024)} \\
    \midrule
    Qual reference: Braun \& Clark (2006) inductive thematic analysis & Qual reference: Braun \& Clark (2017) thematic analysis & Qual reference: Richards \& Hempill (2018) grounded theory and thematic analysis \\
    1. Familiarize with data: ``researcher cleans the data, saves individual files in .txt format, creates interview chunks'' & 1. ``Embed and cluster sentences'' & 1. ``Independent Open Coding, facilitated by on-demand code suggestions from LLMs, yielding initial codes'' \\
    2a. Initial codes: ``LLM identifies initial codes \& quotes from interview chunks, it also provides their descriptions''. 2b. ``LLM reduces duplicate codes, but keeps different quotes'' & 2. ``Generate 15-word summaries of those clusters'' & ``2. Iterative Discussion, focusing on conflict mediation within the coding team, producing a list of agreed-upon code decisions'' \\ 
    3. Searching themes: ``LLM identifies themes and their descriptions by sorting and grouping initial codes and codes descriptions'' & 
    3. ``Cluster the summaries'' & ``3. Codebook Development, where code groups may be formed through LLM-generated suggestions, based on the list of decided codes'' \\
    4. ``Generate short labels of those summary groupings'' & 4. Reviewing themes: ``LLM produces 3 new set of themes, using increased creativity (temp>=0.5). Research reviews for consistent themes for final selection'' &  \\
    5. Defining themes: ``LLM (re)names themes only using the codes, descriptions and quotes. Researcher confirm themes''
    &  &  \\
    \bottomrule
    \end{tabularx}
    \label{tab:llm-assisted-thematic-analysis}
\end{table}

\section{Method}

This Methods section is divided into three subsections. The first subsection describes the data simulation process used to generate the synthetic datasets used in this study. The second subsection describes the central contribution of the paper: the GATOS workflow used to generate the codebooks for qualitative data analysis. Finally, the third subsection describes the evaluation process used to compare the themes generated by the GATOS workflow with the themes and sub-themes embedded in the data simulation process.

\subsection{Data Simulation}
Data for these validation studies came from three simulated datasets. The datasets were simulated to mimic the following three contexts that are encountered in organizational research:
\begin{enumerate}[nosep]
    \item Teammate feedback
    \item Organizational cultures of ethical behavior
    \item Employee perspectives about returning to their offices after the pandemic
\end{enumerate}

These three contexts might be pertinent to organizational research when understanding interpersonal teammate dynamics, how organizations' members perceive the organization's culture of ethical behavior, and the employee perspectives and attitudes toward policy shifts regarding workplace location.  

Our data simulation approach was inspired by common approaches to methods development studies in quantitative research wherein researchers generate synthetic data to test their methods \cite{morris2019using,burton2006design}. The philosophy is that if we control the data generating process then we know what the method \textit{should} identify as the correct answer. For quantitative methods, that might be recovery of parameters used to generate the data. For qualitative methods such as we describe in this paper, the analog could be recovery of themes or sub-themes used to generate the text data.

\begin{figure}[ht]
    \centering
    \includegraphics[width=0.8\textwidth]{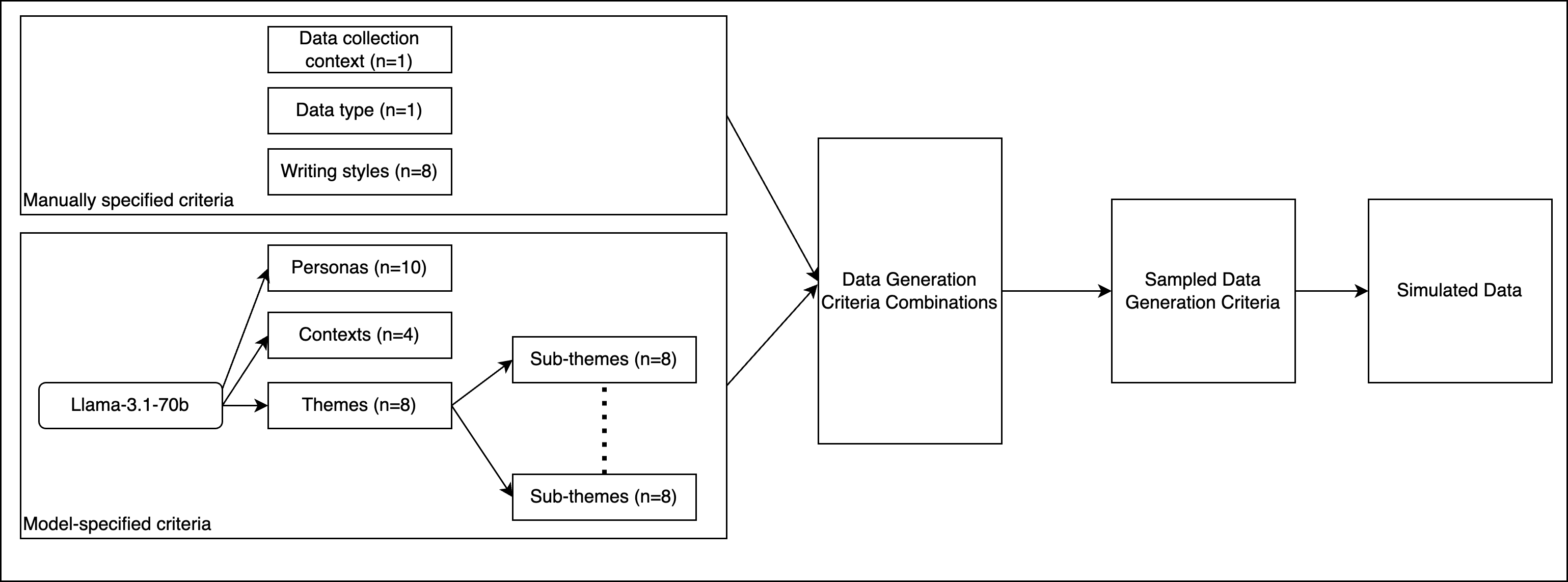}
    \caption{Data Generation Process}
    \label{fig:data-gen-process}
\end{figure}

The objective for generating synthetic data was to generate data that were as realistic as possible to data that one might actually encounter when running studies in organizational research. We achieved that objective in a multi-step process. Step one involved generating backstories and data generation criteria for the model to use. Step two involved using generative text models to actually simulate data according to those generation criteria. There were two categories of criteria used in this process: those generated by a generative text model and those manually specified by the researchers. The model-specified criteria included: personas, contexts, themes, and sub-themes. The manually-specified criteria included: data type, data collection context, writing style, and writing length. The process for data simulation is shown in Figure \ref{fig:data-gen-process}. 

The first step in the data simulation process was to generate personas. We used the Llama-3.1-70b generative text model to generate a set of personas that one might encounter in the context of the specified study. Each persona was a set of attributes that described an individual. For example, a persona might have attributes such as age, occupation, and personality traits. The next step was to generate contexts. For each dataset, we generated four imaginary contexts. Once again, we used Llama-3.1-70b to generate these sets of contexts based on the overall background that the data were supposed to come from. Each context was a description of a situation or environment. For example, a context might describe a workplace like ``Government Contracting Agency''. Next, we generated a list of eight themes that might appear in data collected for the study. Then, for each theme for each dataset we generated a list of eight sub-themes. Again, both of these generation steps used Llama-3.1-70b. We also manually specified writing lengths (e.g., short, medium, long) and writing styles (e.g., professional, casual, sentence fragments) for the data. Finally, we also specified three models that could be used for the data generation process: Wizardlm2-8x22b \cite{xu2023wizardlm}, mistral-nemo-12b, and Llama-3.1-8b \cite{dubey2024llama}.

Specifying this combination of models, personas, contexts, writing styles, writing lengths, themes, and sub-themes led to a combinatorially large number of hundreds of thousands possible data to simulate. Rather than generating millions of simulated responses, we sampled from these possible data generation criteria combinations to generate the actual synthetic datasets used in this study. Specifically, we randomly sampled 18 possible data points for each sub-theme. From a list of 64 sub-themes (8 themes with 8 sub-themes each), this theoretically would result in 1,152 data points for each dataset. In practice, this number was smaller due to some overlap in sub-themes and issues with the data simulation process - some models used for data generation did not follow instructions well, so their outputs were discarded.

A listing of the types of data generation criteria used in this process that were generated by a language model is shown in Table \ref{tab:model-specified-data-gen-criteria}, and a listing of the manually specified data generation criteria is shown in Table \ref{tab:manually-specified-data-gen-criteria}.


\begin{table}[htbp]
    \centering
    \fontsize{10}{12}\selectfont
    \caption{Data Generation Criteria Generated by Text Model}
    \begin{tabularx}{\textwidth}{>{\raggedright\arraybackslash}p{0.2\textwidth} >{\raggedright\arraybackslash}p{0.3\textwidth} >{\raggedright\arraybackslash}X}
    \toprule
    \textbf{Data Generation Criterion} & \textbf{Criterion Description} & \textbf{Example} \\
    \midrule
    Personas & A set of attributes that describe an individual & Union Member \\
    Persona Description & A description of a persona & A 50-year-old suburban resident working in manufacturing. He is a union member who values job security and fair labor practices, and believes that returning to the office should be subject to collective bargaining agreements. This persona may be cautious about returning to an office setting due to concerns about worker rights and protections. \\
    Contexts & A description of a situation or environment & Government agency workplace \\
    Context Description & A description of a context & A bureaucratic setting with strict protocols, hierarchical structures, and a focus on public service. Participants may express concerns about commuting time, office politics, and adapting to new health and safety regulations. \\
    Themes & A list of themes that might appear in a study & Frustration with Lack of Clear Communication \\
    Sub-themes & A list of sub-themes that might appear in a study under that theme; more specific than themes & Insufficient Information About Office Safety Protocols \\
    \bottomrule
    \end{tabularx}
    \label{tab:model-specified-data-gen-criteria}
\end{table}


\begin{table}[htbp]
    \centering
    \fontsize{10}{12}\selectfont
    \caption{Manually Specified Data Generation Criteria}
    \begin{tabularx}{\textwidth}{>{\raggedright\arraybackslash}X>{\raggedright\arraybackslash}X>{\raggedright\arraybackslash}X}
    \toprule
    \textbf{Data Generation Criterion} & \textbf{Criterion Description} & \textbf{Example} \\
    \midrule
    Data Type & The type of data being generated & Written responses \\
    Data Collection Context & The context in which the data were collected & Employee perspectives on returning to work after the pandemic \\
    Writing Style & The style of writing used in the data & Professional \\
    Writing Length & The length of the written responses & Medium (4-5 sentences) \\
    Model & The generative text model used to generate the data & Mistral-nemo-12b \\
    \bottomrule
    \end{tabularx}
    \label{tab:manually-specified-data-gen-criteria}
\end{table}

The following subsections describe the data generation criteria and descriptive statistics for the three generated datasets.

\subsubsection{Simulated Dataset 1: Teammate feedback}

The first synthetic dataset comes from a hypothetical study of teammate feedback in organizations where teamwork is essential. The specified context that the model was given as:

\begin{tcolorbox}[colback=gray!10, colframe=gray!80, title=Context for Teammate Feedback Dataset, boxrule=0.5mm, left=1mm, right=1mm]
\fontsize{10}{12}\selectfont
\textbf{Data Collection Context:} Teammate feedback surveys instructing students to respond to the prompt ``Please provide constructive comments about your fellow teammates as well as yourself. The purpose of these comments is to give you the opportunity to explain how you rated your peers and if there was behavior or experience in particular that influenced you when doing your peer and self evaluations.''
\end{tcolorbox}

\begin{figure}[htbp]
    \centering
    \includegraphics[width=0.4\textwidth]{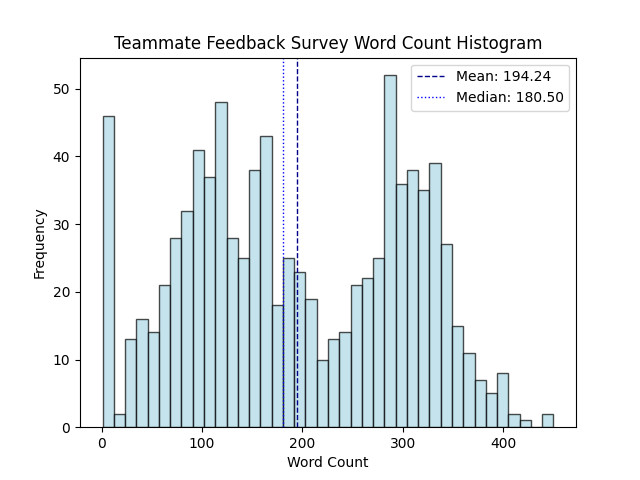}
    \caption{Distribution of Simulated Response Lengths for Teammate Feedback}
    \label{fig:teammate-feedback-response-lengths}
\end{figure}

After removing invalid responses, the simulated teammate feedback dataset consists of 854 written responses to that data collection context prompt. The distribution of length of the written responses is shown in Figure \ref{fig:teammate-feedback-response-lengths}. The average number of words in the responses was 194 and the median was 180. As shown in the figure, the responses demonstrated a bimodal distribution. This distribution was likely a function of the way the data were generated by forcing variety in response length through the writing style and writing length criteria.

An example response from the persona of a 25-year-old ambitious graduate student in this synthetic dataset about teammate feedback is given below:

\begin{tcolorbox}[colback=gray!10, colframe=gray!80, title=Example Simulated Data for Teammate Feedback, boxrule=0.5mm, left=1mm, right=1mm]
    \fontsize{10}{12}\selectfont
    \textbf{Theme:} Leadership and Initiative Taking \\
    \textbf{Sub-theme:} Takes Initiative in Group Discussions \\
    \textbf{Simulated Response:} In reflecting on our group dynamics, I believe I contributed effectively by facilitating note-taking and ensuring all voices were heard. However, I want to highlight the exceptional leadership demonstrated by Alex in driving our group discussions forward. In particular, during our brainstorming sessions for the project proposal and presentation outline, Alex took the initiative to summarize key points, keep us on track, and encourage everyone's input. These instances significantly influenced my peer evaluation, as they not only showcased Alex's strong leadership skills but also fostered a productive environment that led to our group's success.
\end{tcolorbox}

\subsubsection{Simulated Dataset 2: Organizational Cultures of Ethical Behavior}

The second synthetic dataset is from a hypothetical study of organizational cultures of ethical behavior. The context for this imagined study was: 

\begin{tcolorbox}[colback=gray!10, colframe=gray!80, title=Context for Org Culture of Ethical Behavior Synthetic Dataset, boxrule=0.5mm, left=1mm, right=1mm]
\fontsize{10}{12}\selectfont
\textbf{Data Collection Context:} A study asking people working for large companies the following question: ``What factors do you believe affect your organization's culture around ethical behavior?''
\end{tcolorbox}

\begin{figure}[htbp]
    \centering
    \includegraphics[width=0.4\textwidth]{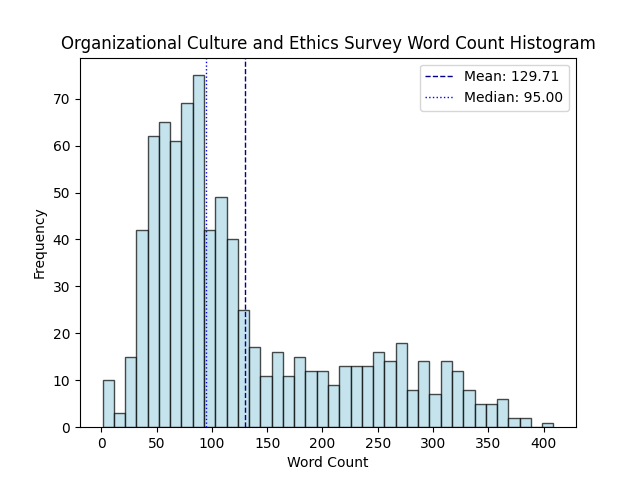}
    \caption{Distribution of Response Lengths for Faculty Perspectives Upon Returning to Work After the Pandemic}
    \label{fig:org-ethics-response-lengths}
\end{figure}

After removing invalid responses, there were 823 responses in this dataset. The distribution of the length of those written responses for this scenario is shown in Figure \ref{fig:org-ethics-response-lengths}. The average number of words in the responses was 129 and the median was 95. The responses demonstrated a long tail distribution, with a few responses being much longer than the rest. This was a deviation from the bimodal distribution of first dataset.

An example response from the persona of a 32-year-old HR specialist in this synthetic dataset about organizational cultures of ethical behavior is given below:

\begin{tcolorbox}[colback=gray!10, colframe=gray!80, title=Example Simulated Data for Organizational Cultures of Ethical Behavior, boxrule=0.5mm, left=1mm, right=1mm]
    \fontsize{10}{12}\selectfont
    \textbf{Theme:} Organizational Size and Complexity \\
    \textbf{Sub-theme:} Geographic Dispersion and Remote Work Challenges \\
    \textbf{Simulated Response:} In my experience, our organization's vast size and global reach have been both a blessing and a curse when it comes to maintaining a strong ethical culture. The sheer number of employees and their diverse backgrounds can make it challenging to ensure everyone adheres to the same ethical standards. Additionally, with more team members working remotely, we've seen a decline in spontaneous conversations that might otherwise catch unethical behavior early on.
\end{tcolorbox}

\subsubsection{Simulated Dataset 3: Employee Perspectives About Returning to Their Workplaces After the Pandemic}

The third synthetic dataset is from a hypothetical study of employee perspectives about returning to their offices after the COVID-19 pandemic. The specified context for this third study was:

\begin{tcolorbox}[colback=gray!10, colframe=gray!80, title=Context for Returning to Office After Pandemic, boxrule=0.5mm, left=1mm, right=1mm]
\textbf{Data Collection Context:} A study of worker perspectives on returning to the office after the pandemic.
\end{tcolorbox}

\begin{figure}[htbp]
    \centering
    \includegraphics[width=0.4\textwidth]{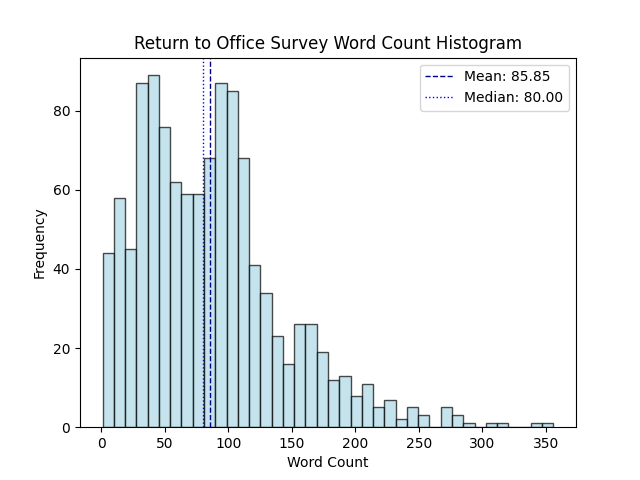}
    \caption{Distribution of Response Lengths for Student Extracurricular Activity Participation}
    \label{fig:employee-return-response-lengths}
\end{figure}

The distribution of the length of the 1,110 written responses for scenario number three is shown in Figure \ref{fig:employee-return-response-lengths}. The average number of words in the responses was 131 and the median was 97. The responses demonstrated a more realistic long tail distribution similar to the second dataset but with fewer responses being much longer than the rest in comparison with dataset number two. This distribution is more similar to our observations from prior work \cite{Katz2024using} where we found that responses to open-ended survey questions often have a long tail distribution.

An example simulated response from the persona of a working mother in this synthetic dataset about perspectives on returning to the office after the pandemic is given below:

\begin{tcolorbox}[colback=gray!10, colframe=gray!80, title=Example Simulated Data for Return to Work Study, boxrule=0.5mm, left=1mm, right=1mm]
    \fontsize{10}{12}\selectfont

    \textbf{Theme:} Appreciation for Remote Work Benefits \\
    \textbf{Sub-theme:} More Comfortable and Personalized Workspace \\
    \textbf{Simulated Response:} I'm really dreading the idea of going back to the office full-time. Working from home has been a game-changer for me - I've been able to create a workspace that is comfortable and personalized to my needs. My home office is set up exactly how I like it, with good lighting, a comfortable chair, and minimal distractions. In contrast, our office can be noisy and chaotic, which makes it hard for me to focus. Plus, the commute is exhausting and takes away from time I could spend with my family. I'm worried that going back to the office will mean sacrificing some of the flexibility and autonomy I've gained while working remotely. I wish our company would consider offering more flexible work arrangements or allowing us to continue working from home at least part-time. That way, I could maintain some of the benefits of remote work while still being able to collaborate with my colleagues in person.
\end{tcolorbox}

Across the three simulated datasets, we believe there is a good representation of the kinds of data that one might encounter in organizational research settings both in terms of topics and lengths on open-ended questions from surveys. With these three synthetic datasets generated, we were able to test the GATOS workflow to see how well the process could identify the themes and sub-themes used to generate the data. The next subsection describes the GATOS workflow, starting first with an overview and then a detailed description of each step in the workflow.

\subsection{GATOS Workflow Overview}
The general task we are trying to solve with this method is simple: identify recurring patterns in text data collected in organizational research settings. Our solution was to combine several NLP tools and techniques with generative text models into a multi-step workflow. In particular, we use an open-source generative text model Mistral-22b-2409 and modern NLP techniques (i.e., text embedding) to enable inductive qualitative data analysis at a large scale. By inductive, we mean a data-driven approach to generate a codebook rather than purely theory-driven. This is the kind of approach one may take when not knowing \textit{a priori} what is discussed in the text. The generated codebook can then be used to label the original text units according to the themes they express. This codebook application step is beyond the scope of the current paper and will be described in future papers. 

We break this inductive codebook generation process down into multiple parts. First, summarize the original text units of analysis. For example, with open-ended survey questions these units of analysis would be the individual written responses from each participant. Second, take all of the atomic summary points and use a text embedding model to generate high-dimensional numeric representations of each summary point. Third, reduce the dimensionality of those text embeddings to a lower dimensional space to enable step four - clustering the embedded summary points. In theory, these clusters should contain semantically similar summary points. For example, summary points such as `frequent emails reminders' and `constantly sent info to help stay on track' could be clustered together since they are about the same idea (i.e., team leader communication about progress). At this point in the process, we have gone from an initial set of raw text units to a smaller set of clusters of semantically similar text units; however, we still do not know what those are necessarily about - only that most of the clustered texts are similar in some way.

Up to now, many of these steps have been explored in prior research. The novelty of this workflow comes in step five. The goals of step five (codebook generation) are two-fold and contend with each other: (1) to generate a code for each cluster and (2) try to avoid generating too many redundant codes. This step is intended to mimic how one might code data in a more traditional qualitative data analysis setting. For example, a human researcher might start by reading their data and generating codes as they encounter new ideas. However, when they see data that fits under a code that already exists in their burgeoning codebook then they would not generate a new code. Instead, they would simply move on to the next data point. At each data point, therefore, the researcher is doing multiple things: identify what is being discussed in the data; think about whether it warrants a code; check whether a code already exists that would describe it; and if not, create that new code and add it to the codebook. 

In step five, we mimic that reflection process with the generative text model and retrieval augmented generation (RAG). In particular, we first take the cluster of text and find the $k$ most similar matches for each summary point in the cluster (usually, $k$ is on the order of 2 to 4) using the embeddings for the summary points, embeddings for the existing codes in the codebook, and calculating their cosine similarities. We then aggregate those nearest neighbors into a set, thereby removing redundancies, to generate a list of unique nearest neighbor codes for that cluster. Consider the scenario where there are six summary points. If there are six summary points in a cluster, we would find $6 * k$ nearest neighbor codes for that cluster, though this list may have multiple copies of the same codes. Staying with the same example as above, if the six summary points are all about the team leader sending email reminders, then we try to find the existing codes in the codebook that are most similar to this idea, but that may only result in two or three unique existing codes in the codebook that might be similar to those summary points. Those unique codes are then included in the prompt to the generative text model when it is deciding whether to generate a new code for that cluster.

At a philosophical level, at this point in the process, it is still an open question whether those `most similar' codes actually do capture the idea(s) expressed in that cluster. To investigate this question, we instruct a generative text model to look at the cluster of ($n$) summary points, the existing codes in the codebook (as represented by the $\leq n * k$ nearest neighbor codes), and decide whether or not a new code is needed based on whether the existing codes in the codebook provide sufficient thematic coverage of the cluster of summary points. If the model decides that a new code (or codes) is (are) needed, then it generates a new code and definition for that code, which are added to the codebook. If the model decides that a new code is not needed, then the process simply moves on to the next cluster of data points. The full prompt for this code consideration step is given in the Appendix \ref{subsec:cb-create-prompt}. 

\begin{figure}[htbp]
    \centering
    \includegraphics[width=0.8\textwidth]{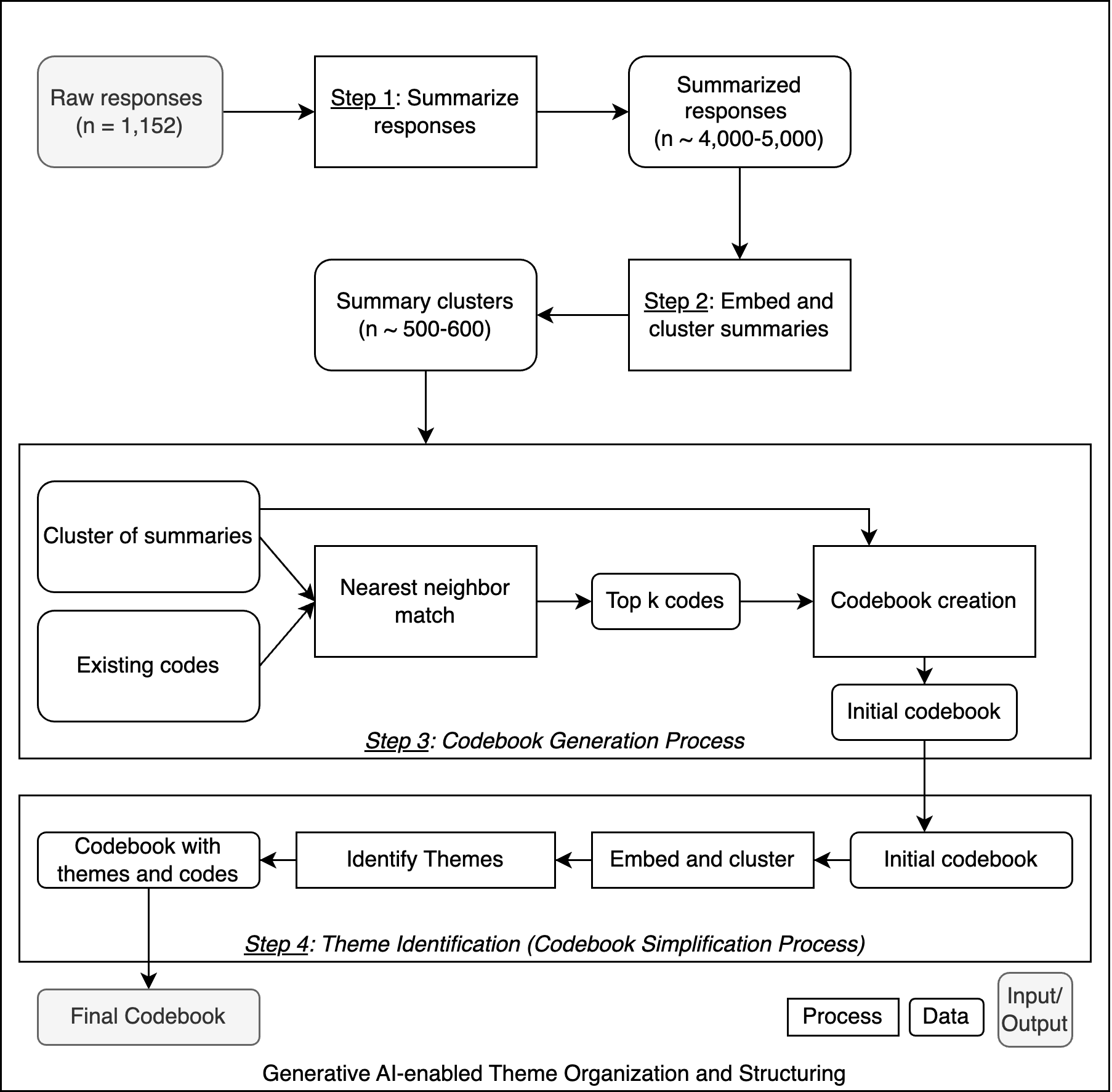}
    \caption{Workflow for the method}
    \label{fig:method-overview}
\end{figure}

By the end of step five, we have gone through each cluster of summary points and either identified a new code or decided that no new code was needed, thereby balancing the goals of identifying the recurring patterns in the summary points while also not generating too many redundant codes. In practice, there are still many near redundancies that exist, so we proceed to a final step in which we cluster these newly generated codes and prompt a model to identify the distinct themes. The process therefore culminates with a list of themes and codes belonging to those themes. This entire workflow is shown in Figure \ref{fig:method-overview}. We call this workflow the Generative AI-enabled Theme Organization and Structuring (GATOS) method because it is designed to mimic parts of thematic analysis while also being distinct from actual traditional thematic analysis.

For evaluation purposes in this study, the final step in our work here was to compare the themes generated through this process with the ($8$) themes and ($8 * 8$) sub-themes used to simulate the original data. The details of each step in the workflow, along with procedures to evaluate our workflow, are described in more detail below.

\subsection{GATOS Workflow in Detail}
This section presents the steps used in the GATOS workflow to generate a codebook inductively from raw data. At a high level, the goal is to mimic steps a human coder might take when conducting thematic analysis by first reading the raw data, summarizing them in multiple summary points, generative codes that capture recurring patterns, and then identifying themes among those codes. Ideally, new codes are generated when a key idea in the data is not yet represented by a code in the existing codebook. In practice, we instruct the generative text model to read a group of $n$ summaries, compare it with $\leq n * k$ of the nearest neighbor matches for the codes in the codebook, and then decide whether or not a new code should be generated to describe the idea(s) expressed in that cluster of summary points. The details of the workflow steps are described in the following subsections.

\subsubsection{Step 1: Summarize the Original Data}
The first step in the GATOS workflow is to summarize the original data into distinct ideas. This step is necessary because the raw data may contain multiple ideas in a single response. For example, a participant may write a long response that contains multiple ideas about their perspective of their organization's culture of ethics. The goal of this step is to extract the key ideas from the raw data to make them easier to analyze in the subsequent steps. One can use a relatively small language model (e.g., 7 to 14 billion parameters) for this step. Here, we used the mistral-nemo-12b model for this information extraction task because it is small enough to run quickly but performant enough not to miss parts of the original data to summarize (along with being an open-source model and released under an Apache-2.0 license). The prompt used for summarization is given in the Appendix \ref{subsec:initial-summarization-prompt}. An example of the input and output for this information extraction is shown in table \ref{tab:info-extract-example}.

\subsubsection{Step 2: Clustering Semantically Similar Ideas}
The second step in the workflow is to identify semantically similar summary points from step one. The goal of this step is to group together the instances when the same idea is expressed in different ways in the data. If we can do this grouping well, then we should be able to proceed through the clusters one-at-a-time to identify recurring patterns more easily than if there were several discrete ideas covered in each cluster. To accomplish step 2, we first embed all of the summary ideas using a text embedding model. In our case here we used the `mxbai' model \cite{li2023angle} because it is a lightweight performant model according to the MTEB leaderboard \cite{muennighoff2022mteb} released under an Apache-2.0 license. The next step is to reduce the dimensionality of these 1,024-dimensional embeddings because clustering in this high-dimensional space might suffer from the curse of dimensionality \cite{assent2012clustering}, where all points tend to be far from each other in high-dimensional spaces. We use principal component analysis (PCA) plus uniform manifold approximation and projection (UMAP) \cite{mcinnes2018umap} to reduce the dimensionality of the embeddings to a lower-dimensional space. We first use PCA to reduce from 1,024 dimensions to $D$ dimensions, where $D$ is dynamically found by identifying whichever dimension number retains 90\% of the variance in the data. In practice, we tended to find $100 \leq D \leq 120$. From this intermediate embedding space, we then used UMAP to reduce down to five dimensions. We use UMAP here to try and preserve the global structure of the data manifold and improve clustering results. From this low-dimensional space, we then use agglomerative clustering with Euclidean distance to cluster the summary points. We use agglomerative clustering rather than K-means because our initial testing suggested that more homogeneous clusters were identified with agglomerative clustering. The output of this step is a set of clusters of semantically similar summary points.

\subsubsection{Step 3.1: Create Set of Speculative Starter Codes}
The next step in the GATOS workflow is to create the actual codebook by iteratively reading the clusters of summary points and deciding whether to generate a new code or not. To start the process, however, we prompt a generative text model to create 20 hypothetical codes that one might expect to appear in a study of whatever topic we simulated. For example, for the teammate feedback study, we prompted the model to generate 20 hypothetical codes one would expect to describe some data collected in a study of teammate feedback. These initial synthetic codes are used to help the process get started by providing the process with some initial codes to consider from the nearest neighbor matching when generating new codes. The specific prompt is given in the Appendix \ref{subsec:cb-create-prompt}.

\subsubsection{Step 3.2: Inductive Codebook Generation}

With the clusters of summary points and speculative starter codes generated, we can now begin the iterative inductive codebook generation. We accomplish this by embedding each summary point that belonged to a cluster. We once again used the mxbai embedding model as in Step 2. We then embed all entries in the codebook (at the start, this is 20 codes, but as the codebook grows this number also grows). 

After embedding the extracted information and the codes from the growing codebook, we find the $k$ nearest neighbors codes for each of the extracted ideas in the cluster. We use cosine similarity for this matching. Cosine similarity is a measure of similarity between two non-zero vectors belonging to an inner product space. The similarity score is defined to equal the cosine of the angle between them, which is also the same as the inner product of the same vectors when normalized to both have unit length. In plain language, cosine similarity measures similarity between two vectors, e.g., embedding vectors. We use the cosine similarity to find the $k$ nearest neighbors for each of the extracted ideas. In the present study, we set $k=5$. We chose 5 because to balance between having enough potentially relevant codes in the codebook for the generative text model to check without giving the model \textit{all} of the codes. Giving the model all of the codes might be a bad idea because it could distract or overwhelm the model's attention. The output of this neighbor matching step is a set of $k$ nearest neighbors for each of the extracted ideas. We use these as part of the prompt to the generative text model to decide whether to generate a new code or not based on the extracted idea and the extant codes in the codebook.  

Simply put, the procedure is as follows. First, embed the new cluster of summary points that the model is going to analyze. Next, embed all the entries in the codebook. Then, calculate cosine similarity between the embedding for the new text and the codes in the codebook to find the $k$ nearest codes in the codebook for the new piece of text. In our case, we set $k$ as 5. Remove duplicate codes from this set. Finally, include only this small subset of the codes in the prompt to the generative text model when instructing it to consider whether or not to generate a new code based on the new cluster it is reading and the subset of the codebook it can reference.

\subsubsection{Using Generative Text Model to Decide Whether to Generate a New Code}

The next step in our process runs $C$ many times, where $C$ is the number of clusters from step two. 
As mentioned before, at a high level we instruct the generative text model to read the summary points in a single cluster, read the nearest neighbor codes from the codebook that \textit{might} describe the ideas in that cluster, and then decide whether to generate a new code or not. We used Mistral-22b-2409 for this step because it is sufficiently large to be able to exhibit reasoning steps while being small enough to run quickly on common consumer hardware. The prompt used for this step is given in the Appendix \ref{subsec:cb-ind-gen-prompt}.

There are six specific steps that the model is instructed to complete as part of this decision-making process. First, the model is instructed to read the existing codebook. Second, the model is instructed to read the summary data points in cluster. Next, the model is instructed to try to use one or more of the existing codes to describe the summaries in the cluster. Fourth, the model is instructed to create a new code if needed. The model then must evaluate whether the suggested new code adheres to three evaluation criteria: parsimony, abstraction level, and non-redundancy. Finally, the model is instructed to make a final recommendation about whether to create a new code or not.

This workflow sounds complicated, but the actual philosophy is similar to what a traditional qualitative research might do when creating a codebook inductively: start reading data, create a code, and add that to the codebook. Then, when reading the next piece of data, consider the existing codes and see if the new data can be described by one of those existing codes in the codebook. If the new data point already has a code that describes it, then there is no need to generate a new code, so proceed to the next piece of data. On the other hand, if the new data does not already have a code that describe it, then create a new code. We repeat this process for each of the $n$ pieces of data (i.e., text) in the dataset. 



\subsection{Step 4: Codebook Simplification Through Theme Identification}

The final step in the method is to simplify the codebook by trying to identify themes. This step is necessary because the generative text model may generate redundant codes in the preceding step and codes that belong together at a more abstract level. To address these issues, we used a step in which we clustered similar codes together and gave those to the model along with the instructions to identify higher-level themes in the clusters of codes where possible. The specific prompt used in this step is given in the Appendix \ref{subsec:cb-theme-identification-prompt}. The final output of the GATOS workflow is a set of themes and codes that should describe the common semantic patterns in the data.

\section{Results}
In this results section, we will present the results of the simulation study across the three synthetic datasets. For each dataset, we compare the codebooks generated by the GATOS workflow with the specified codebooks used to generated the original data. Before reporting results from each case, we first give examples of input and output for each of the main prompts used in this process.

\subsection{Example Input and Output for Each Generative Step in GATOS Workflow}

\subsubsection{Information Extraction with Generative Text Model}

An example of the input and output from the information extraction step is given in table \ref{tab:info-extract-example}. We ran this step for each entry in the datasets to extract the key ideas from each response, which makes the data more manageable for the next steps in the method. As shown here, the input is a medium-length simulated response from the return-to-office dataset, and the output is a five-point list of key ideas extracted from that response. The specified theme and sub-theme for the original synthesized data were ``Expectations for Hybrid Work Arrangements'' and ``Fears about Negative Impact on Career Advancement'', respectively. Despite being given a single theme and sub-theme to discuss, the generative model included more than just those themes and sub-themes in the simulated response. This can be seen in the summary points extracted from the response, which include a mix of the specified theme and sub-theme as well as additional information that was not part of the original specified criteria (but was in the simulated data as a byproduct of the simulation).

\begin{table}[htbp]
    \centering
    \fontsize{10}{12}\selectfont
    \caption{Example of Information Extraction from Return to Work Synthetic Dataset}
    \begin{tabularx}{\textwidth}{>{\raggedright\arraybackslash}X>{\raggedright\arraybackslash}X}
    \toprule
    \textbf{Input} & \textbf{Output} \\
    \midrule
    ``I'm really excited about returning to the office, but I have to admit that I'm worried about how hybrid work arrangements will affect my career advancement. If we're not in the office regularly, I'll miss out on watercooler conversations with clients and colleagues, which can lead to new business opportunities. I've seen it happen before - when people are working remotely, they tend to get stuck in their own little bubbles and don't have as many chances to connect with others. I'm hoping that our company will find a way to balance flexibility with face-to-face interaction, so we can all continue to grow and succeed together.'' & ``My expert summary:

    1. Excited about returning to the office
    
    2. Concerned about career advancement in hybrid work arrangements
    
    3. Fears missing out on networking opportunities with clients and colleagues
    
    4. Believes remote work can lead to isolation and reduced connections
    
    5. Hopes company balances flexibility with face-to-face interaction for collective growth'' \\
    \bottomrule
    \end{tabularx}
    \label{tab:info-extract-example}
\end{table}

\begin{figure}[htbp]
    \centering
    \includegraphics[width=0.4\textwidth]{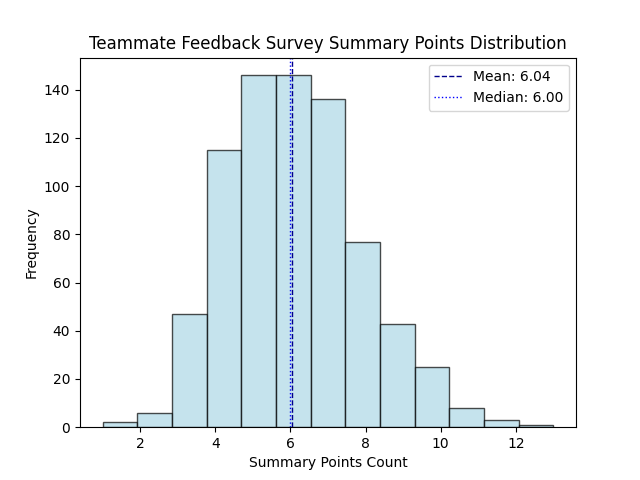}
    \caption{Distribution of Summary Points Per Response for Teammate Feedback}
    \label{fig:teammate-feedback-summary-point-distribution}
\end{figure}

\begin{figure}[htbp]
    \centering
    \includegraphics[width=0.4\textwidth]{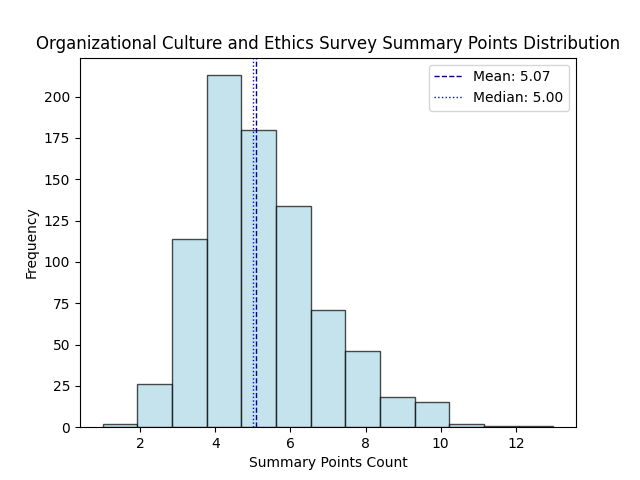}
    \caption{Distribution of Summary Points Per Response for Perspectives on Organizational Cultures of Ethical Behavior}
    \label{fig:org-ethics-summary-point-distribution}
\end{figure}

\begin{figure}[htbp]
    \centering
    \includegraphics[width=0.4\textwidth]{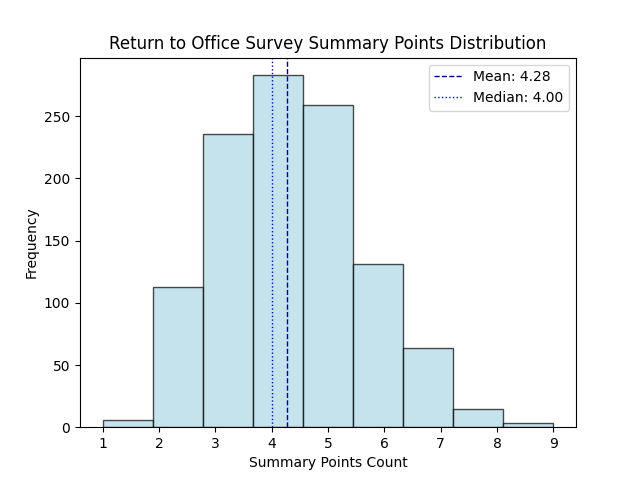}
    \caption{Distribution of Summary Points Per Response for Returning to Work After the Pandemic}
    \label{fig:return-to-work-summary-point-distribution}
\end{figure}

The distributions of the number of summary points extracted for the teammate feedback dataset is shown in Figure \ref{fig:teammate-feedback-summary-point-distribution}, for the organizational culture of ethics synthetic dataset in Figure \ref{fig:org-ethics-summary-point-distribution}, and for the return to workplace dataset in Figure \ref{fig:return-to-work-summary-point-distribution}. Coincidentally, the average number of summary points extracted from reach response for the three datasets was 6.04, 5.07, and 4.28, respectively. These values give readers a sense of the amount of information extracted from each response and how the original data (i.e., full response) can expand 4x-6x as it is broken into summary points for the next step in the process. Additionally, the distributions of these summary points more closely resembled normal distributions, which was a notable contrast compared to the long-tailed and bimodal distributions of the length of the original synthesized data.

\subsubsection{Inductive Codebook Generation}

After extracting information from each response to generate summary points, we then clustered the summary points and prompted a generative text model to iteratively consider whether or not to create a new code to describe that cluster. In the codebook creation prompt, the codes in the existing codebook XML tag in the prompt were the nearest codes to the new cluster of summary points. At this point in the GATOS workflow, there are two types of possible outcomes from the generative text model: (1) a new code (or several) is generated and added to the codebook or (2) no new code is generated. The following three gray boxes present the input and output for the inductive codebook generation step where a code is created when analyzing the return to work after the pandemic dataset. Following this example where a code is suggested we also provide an example where the recommendation is to not create a new code. This first box shows the input summary points for the cluster.

\begin{tcolorbox}[title=Example Cluster Summary Point Members for Return to Work After the Pandemic Dataset]
    \fontsize{10}{12}\selectfont
    \textbf{Input Summary Points:} \begin{enumerate}[nosep]
        \item requires seamless video conferencing for communication
        \item desires user-friendly video conferencing tools
        \item up-to-date hardware and software solutions required for smooth transitions
        \item requires robust infrastructure for seamless video conferencing and secure VPN access at home
        \item hopes for reliable video conferencing tools (Zoom, Google Meet)
        \item wants seamless communication platforms (Slack, Microsoft Teams)
        \item specific technological needs: high-quality video conferencing, secure VPNs, cloud-based software applications
        \item request for reliable video conferencing tools at home
        \item reliable video conferencing and high-speed internet are necessary
        \item unfamiliar with potential new collaboration tools
    \end{enumerate}
\end{tcolorbox}

We see that many of the summary points in this cluster are about the need for reliable video conferencing tools and secure VPN access for remote work. The nearest neighbor codes for this cluster are shown in the following gray box.

\begin{tcolorbox}[title=Nearest Codes for Cluster in Return to Work After the Pandemic Dataset]
    \fontsize{10}{12}\selectfont
    \textbf{Nearest Codes:} \begin{enumerate}[nosep]
        \item adaptability to change
        \item social interaction needs
        \item adaptability to sudden changes
        \item productivity at home
        \item anticipation of social interaction
        \item training and support needs
        \item information clarity frustration
        \item technological challenges
        \item flexible scheduling and remote work options
        \item navigating unpredictable changes
    \end{enumerate}
\end{tcolorbox}

These nearest neighbor codes discuss changes, productivity at home, social interaction, and technological challenges. From these summary points and nearest neighbor codes, the model generated the following reasoning and concluding code recommendation.

\begin{tcolorbox}[title=Inductive Codebook Generation Example Output for Return to Work After the Pandemic Dataset]
    \fontsize{10}{12}\selectfont
    \textbf{Output:} My expert analysis:

    Step 1 (codebook examination)

    - The existing codebook includes codes related to technology and communication such as "technological challenges" and "flexible scheduling and remote work options".
    
    Step 2 (current data examination)

    - The main theme in the written response is the need for reliable, user-friendly video conferencing tools and robust infrastructure for seamless communication.
    
    Step 3 (analysis part 1)

    - The existing code "technological challenges" partially captures the issues related to video conferencing tools but does not fully encompass the specific needs mentioned in the response.

    - The code "flexible scheduling and remote work options" touches on the need for communication platforms but is more focused on scheduling flexibility rather than the technical aspects of communication tools.
    
    Step 4 (analysis part 2)

    - To better capture the specific technological needs expressed in the written response, a new code could be created to encapsulate these requirements.

    - This new code would focus on the need for reliable and user-friendly video conferencing tools, secure VPNs, and high-speed internet.
    
    Step 5 (reflection on planned suggestions)

    - The creation of this new code aligns with the evaluation criteria as it minimizes redundancy and provides a more precise description of the technological needs mentioned in the response.

    - This code is essential for capturing the nuances of the technological challenges faced by individuals working from home.
    
    My logical recommendation:

    Code: reliable video conferencing tools

    Definition: Need for user-friendly, reliable video conferencing tools (e.g., Zoom, Google Meet), secure VPNs, and high-speed internet to facilitate seamless communication and collaboration.
\end{tcolorbox}

As shown in this example, the model proceeds through each of the instructed steps, starting with a consideration of the existing codes in the codebook. The model then reads the summary points in the cluster and identifies the main theme as the need for reliable video conferencing tools and secure VPN access. The model then considers the existing codes in the codebook and finds that the code ``technological challenges'' partially captures the main theme but does not fully encompass the specific needs mentioned in the response. After more reasoning steps, the model decides that a new code is needed to capture the specific technological needs expressed in the written response. The code suggested was ``reliable video conferencing tools'', which matches some of the main ideas in the summary points yet was not already covered in any existing codes in the codebook. Thus, the generative text model displayed a step-by-step reasoning sequence that culminated in a new code being added to the codebook. In contrast with this example, the next example shows an instance where the model decided not to create a new code. We again use the return to work after the pandemic dataset for this example.

\begin{tcolorbox}[title=Example Cluster Summary Point Members for Return to Work After the Pandemic Dataset (No code)]
    \fontsize{10}{12}\selectfont
    \textbf{Input Summary Points:} 1. open to hybrid work arrangements for balancing work and family commitments

    2. hope company takes concerns seriously and works together on beneficial hybrid arrangements
    
    3. need for hybrid work arrangements that accommodate various needs (e.g., caring for elderly parents or young kids)
    
    4. prefers flexible, hybrid work arrangement for balancing career and family responsibilities
    
    5. hopes for hybrid work model to balance family and job responsibilities
\end{tcolorbox}

This cluster of summary points is about the need for hybrid work arrangements that accommodate various personal needs, such as caring for elderly parents or young kids. The nearest neighbor codes for this cluster are shown in the following gray box.


\begin{tcolorbox}[title=Nearest Codes for Cluster in Return to Work After the Pandemic Dataset (No code)]
    \fontsize{10}{12}\selectfont
    \textbf{Nearest Codes:} 1. work-life balance

    2. team integration anxiety

    3. flexible work hours

    4. hybrid work flexibility

    5. single parent work life balance

    6. job security fears

    7. childcare arrangements
\end{tcolorbox}

The nearest neighbor codes for this cluster of summary points centered around work-life balance, hybrid work flexibility, and family responsibilities. From these summary points and nearest neighbor codes, the LLM generated the following reasoning and recommendation to forego creating a new code.

\begin{tcolorbox}[title=Inductive Codebook Generation Example Output for Return to Work After the Pandemic Dataset (No code)]
    \fontsize{10}{12}\selectfont
    \textbf{Output:} My expert analysis:

    Step 1 (codebook examination)

    - The existing codebook includes codes related to work-life balance, hybrid work flexibility, and family responsibilities.
    
    Step 2 (current data examination)
    
    - The main theme in the written response is the desire for flexible, hybrid work arrangements that accommodate various personal needs, including caring for elderly parents or young kids.
    
    Step 3 (analysis part 1)
    
    - The existing codes "hybrid work flexibility" and "work-life balance" partially capture the main theme but do not fully encompass the specific need for accommodating diverse family responsibilities.
    
    Step 4 (analysis part 2)
    
    - No new codes needed: The existing code "hybrid work flexibility" can be expanded to include the nuance of accommodating various personal needs, thus avoiding unnecessary code creation and maintaining parsimony in the codebook.
    
    Step 5 (reflection on planned suggestions)
    
    - Reflecting on the evaluation criteria, the recommendation maintains the integrity of the existing codes while ensuring that the specific need for diverse family responsibilities is captured without creating redundant or overly specific codes.
    
    My logical recommendation:
    
    No new codes needed

\end{tcolorbox}

In this example, the model decided that no new code was needed because the existing codes in the codebook were sufficient to describe the ideas expressed in the summary points. This response seems plausible because the code ``hybrid work flexibility'' sufficiently covers the central idea in the cluster of summary points, namely the need for flexible, hybrid work arrangements that accommodate various personal needs. This example illustrates the central point of this entire process: to balance the need for new codes to capture the nuances of the data with the need to avoid generating redundant codes. Moreover, this response illustrates how the model can make a reasoned decision that might be similar to what a human coder would do when creating a codebook inductively.

\subsubsection{Theme Creation}

An example of the input cluster of codes and output suggested themes from the theme creation step are given in table \ref{tab:cb-theme-identification-example}. The input codes contained some redundancies related to flexible work arrangements and work hours, so the generative text model identified those as part of the theme ``Flexibility in Work Arrangements''. The model also identified themes related to uncertainty in work schedules and breaks, so those were combined into the theme ``Uncertainty in Work Schedules''. This example also illustrates how individual codes could also persist in the theme identification process as in the case of the code ``Control Over Work Hours''. The example also illustrates how there is a thin line between a code and a theme, as the code ``Control Over Work Hours'' could be seen as a theme in its own right, or it could have been subsumed into the theme ``Flexibility in Work Arrangements''. 

\begin{table}[htbp]
    \centering
    \fontsize{10}{12}\selectfont
    \caption{Example of Theme Identification from Initial Codebook}
    \begin{tabularx}{\textwidth}{>{\raggedright\arraybackslash}X>{\raggedright\arraybackslash}X}
    \toprule
    \textbf{Input Codes} & \textbf{Output Themes} \\
    \midrule
    \begin{enumerate}[nosep]
        \item Flexible work arrangements for wellbeing
        \item Flexible work hours
        \item In office work frequency and day uncertainty
        \item Uncertainty about remote work frequency and future schedules
        \item Flexible work arrangements for caregivers
        \item Individualized flexible work arrangements
        \item Supportive inclusive work environment
        \item Resistance to traditional office hours
        \item Staggered work hours/shifts
        \item Clear workday end time and personal time protection guidelines
        \item Break and lunch dynamics uncertainty
        \item Control over work hours
    \end{enumerate}
    &
    \begin{enumerate}[nosep]
        \item Flexibility in Work Arrangements
        \item Uncertainty in Work Schedules
        \item Supportive Work Environment
        \item Clear Guidelines and Boundaries
        \item Control Over Work Hours
    \end{enumerate}
    \\
    \bottomrule
    \end{tabularx}
    \label{tab:cb-theme-identification-example}
\end{table}

\subsection{Codebook Creation Rates}
\begin{figure}[htbp]
    \centering
    \includegraphics[width=0.5\textwidth]{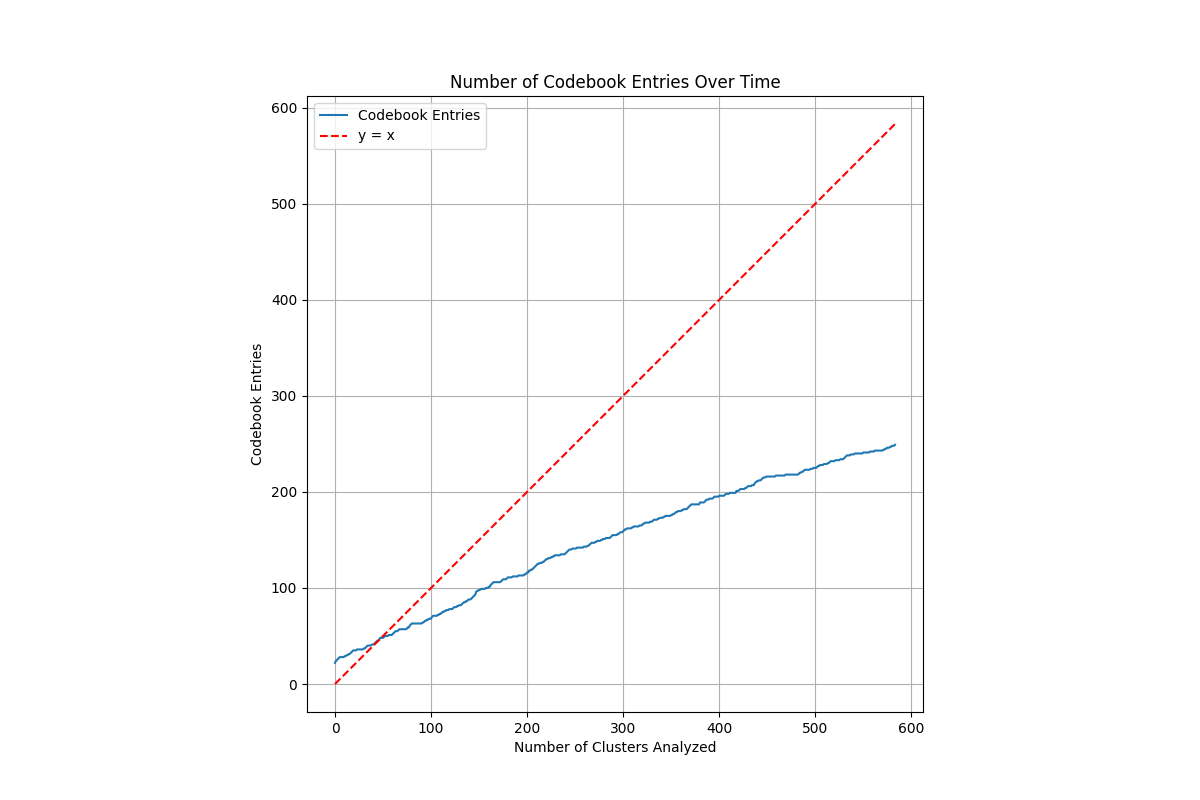}
    \caption{Codes Generated vs Clusters Analyzed for Teammate Feedback}
    \label{fig:teammate-feedback-codes-vs-cluster}
\end{figure}

\begin{figure}[htbp]
    \centering
    \includegraphics[width=0.5\textwidth]{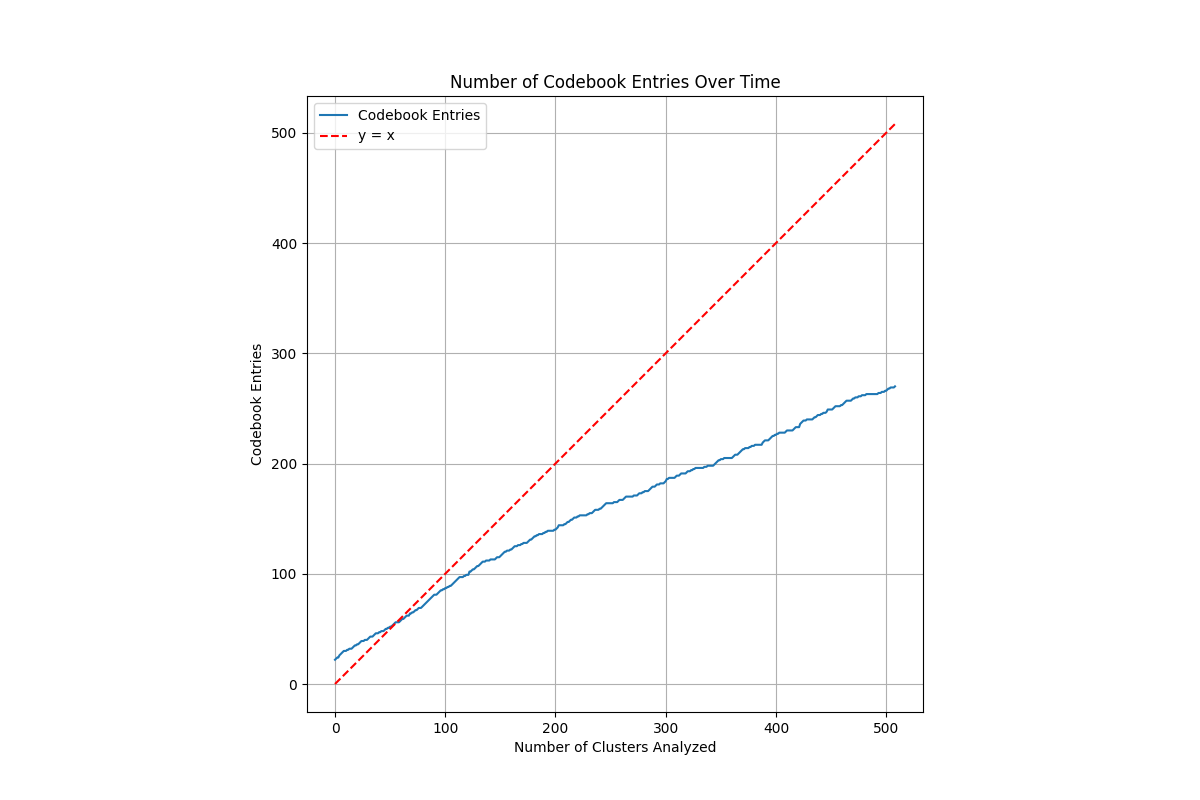}
    \caption{Codes Generated vs Clusters Analyzed for Organizational Culture of Ethics}
    \label{fig:org-culture-codes-vs-cluster}
\end{figure}

\begin{figure}[htbp]
    \centering
    \includegraphics[width=0.5\textwidth]{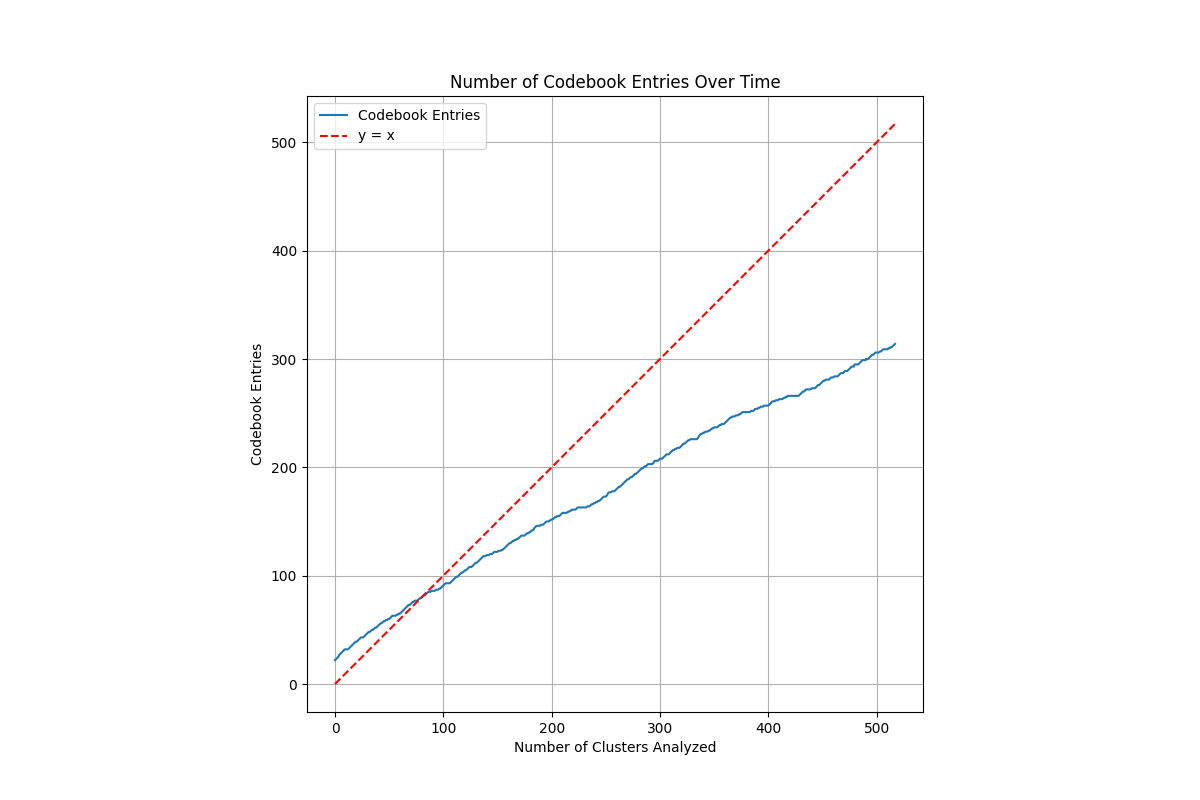}
    \caption{Codes Generated vs Clusters Analyzed for Returning to Work After the Pandemic}
    \label{fig:office-return-codes-vs-cluster}
\end{figure}

One important objective of the GATOS workflow is to generate codes that capture the ideas in the data while avoiding redundancy in the codes. To track this, we noted the rate of code creation over time as each cluster was scrutinized by the generative text model. In the worst-case scenario, the model would create one (or more) new codes with each cluster it encountered. This would be unideal because there would almost certainly be thematically redundant clusters, which would lead to redundant codes being created and defeat the purpose of the codebook checking step. Instead, if the model actually were checking the codes in the codebook to decide to create a new code or not, we would expect to see the model generate fewer new codes over time as it encounters more clusters and the codebook becomes saturated. This would suggest that the model is finding existing codes in the codebook that can describe the ideas in the new clusters. The rate of code creation for the teammate feedback is shown in Figure \ref{fig:teammate-feedback-codes-vs-cluster}, for the organizational culture of ethics dataset in Figure \ref{fig:org-culture-codes-vs-cluster}, and for the return to workplace dataset in Figure \ref{fig:office-return-codes-vs-cluster}.

As the figures show, the GATOS workflow met the minimal success criterion of not always generating a new code for each cluster. This can be seen by the deviation of the blue line from the red 45-degree line in the figures. The blue line represents the cumulative number of new codes generated by the model, while the 45-degree line represents the trajectory one would expect to see the blue line follow if there were a new code created for each cluster. The blue line bending below the red line is a quick way to see that sometimes the model does not create a new code for a cluster. The figures also show how many overall codes are initially created for the three datasets. The summary table of the starting number of simulated data points, summary points, codes, and themes for each dataset is shown in Table \ref{tab:gatos-summary-table}.

\begin{table}[htbp]
    \centering
    \fontsize{10}{12}\selectfont
    \caption{Summary of GATOS Workflow}
    \begin{tabularx}{\textwidth}{
    >{\raggedright\arraybackslash}X
    >{\raggedright\arraybackslash}X
    >{\raggedright\arraybackslash}X
    >{\raggedright\arraybackslash}X
    >{\raggedright\arraybackslash}X
    }
    \toprule
    \textbf{Dataset} & \textbf{Data Points\textsuperscript{1}} & \textbf{Summary Points} & \textbf{Codes\textsuperscript{2}} & \textbf{Themes} \\
    \midrule
    Teammate Feedback & 854 & 4,563 & 249 & 89 \\
    Organizational Culture of Ethics & 823 & 4,176 & 246 & 75 \\
    Return to Workplace & 1,110 & 4,748 & 314 & 110 \\
    \midrule
    \multicolumn{5}{l}{\footnotesize \textsuperscript{1} Some observations dropped due to parsing errors or duplication} \\
    \multicolumn{5}{l}{\footnotesize \textsuperscript{2} Number of codes includes original 20 codes generated by model to initiate process} \\
    \bottomrule
    \end{tabularx}
    \label{tab:gatos-summary-table}
\end{table}

For the teammate feedback dataset, from an initial 854 simulated responses, the workflow generated 249 codes and 89 themes. For the organizational culture of ethics dataset, from an initial 823 simulated responses to the question of factors influencing their organization's culture of ethical behavior, the workflow generated 246 codes and 75 themes. Finally, for the simulated dataset in which respondents were asked about their expectations for returning to the workplace after the pandemic, the workflow generated 314 codes and 110 themes from an initial 1,110 simulated responses. In each case, there is approximately a 10:1 ratio of simulated responses to themes generated by the GATOS workflow. It is an open question whether this is a coincidence or a generalizable pattern. At this point in the process, it is also an open question whether the themes generated by the GATOS workflow matched the sub-themes used to generate the original synthetic data. We address this question in the next section.

\subsection{Evaluation: Comparing GATOS-generated Themes with Original Sub-Themes}
In this section we present the answer to the central question of this paper: \textit{how well do the codebooks generated through the GATOS workflow capture the themes and sub-themes used to generate the original synthetic data?}. We present the answer to this question for each of the three datasets in their respective subsections. In each of those subsections, we present examples where the workflow nearly matches the original themes one-to-one and examples where the workflow only partially captures the original sub-theme. It is important to note that we are only presenting the comparisons for the GATOS workflow-generated themes and the original sub-themes. We do not present the comparisons for the individual codes generated by the GATOS workflow because the codes are not the final output of the workflow; however, we did find in our testing that those codes always matched an original sub-theme perfectly. One might expect that result given the number of codes generated by the workflow (in the range of 246 to 314, as shown in table \ref{tab:gatos-summary-table}) and the number of sub-themes used to generate the original synthetic data.

\subsubsection{Synthetic Dataset 1: Teammate Feedback}
There were 60 sub-themes (after removing a few redundant sub-themes) used to generate the original synthetic data for the teammate feedback dataset. The codebook generated by the GATOS workflow contained 249 codes and 89 themes. An example comparison of the codebook generated by the GATOS workflow with the original sub-themes is shown in Table \ref{tab:comparison-teammate-feedback-best}. In this table, we have selected the best examples of how the GATOS workflow themes aligned with the sub-themes in the original synthetic data. There were many examples where the GATOS workflow-generated themes closely matched the original sub-themes. For example, the original sub-theme ``effective use of feedback mechanisms'' was most closely matched with the GATOS workflow theme ``feedback mechanism effectiveness''. Another near-perfect match was for the sub-theme ``embracing feedback as a learning opportunity'', which was matched with the GATOS workflow theme ``openness to feedback''. The latter example demonstrates how the GATOS workflow can capture the semantic meaning of a sub-theme without using the exact same words (i.e., openness and embracing).

\begin{table}[htbp]
    \centering
    \fontsize{10}{12}\selectfont
    \caption{Best Examples of Comparison of Codebooks for Teammate Feedback Synthetic Dataset}
    \begin{tabularx}{\textwidth}{XX}
    \toprule
    \textbf{Original Synthetic Sub-Theme} & \textbf{GATOS Workflow Method Closest Match} \\
    \midrule
    effective use of feedback mechanisms & feedback mechanism effectiveness \\
    empathy and understanding in communication & empathetic communication \\
    leadership and initiative taking & leadership and initiative \\
    effective communication styles & effective communication \\
    embracing feedback as a learning opportunity & openness to feedback \\
    \bottomrule
    \end{tabularx}
    \label{tab:comparison-teammate-feedback-best}
    \end{table}

To avoid the temptation to cherry-pick the data, we also present the worst matches or unmatched themes for the teammate feedback synthetic dataset in Table \ref{tab:comparison-teammate-feedback-worst}. In this table, we show the instances where there was no clear theme that perfectly aligned with a sub-theme used in the original data generation step. For example, the simulated sub-theme ``setting realistic goals for self-improvement'' was only matched by ``personal growth'' and ``self-reflection'' in the GATOS-generated themes. Those themes are missing an aspect of goal setting that was present in the original sub-theme. The worst match of these poor matches was for the sub-theme ``avoiding blame'', which was matched with the themes ``personal bias management'' and ``negative behaviors and their impact''. While these themes might indirectly relate to avoiding blame, they miss the essence of blame avoidance. It should be noted that the original codes generated by the GATOS workflow (before they were combined into themes) did indeed have a closer match of ``dismissive behavior'', though this, too, misses the aspect of blame. All other 56 sub-themes had at least one good match with the GATOS workflow-generated themes. 


\begin{table}[htbp]
    \centering
    \fontsize{10}{12}\selectfont
    \caption{Worst Comparisons of Codebooks for Teammate Feedback Synthetic Dataset}
    \begin{tabularx}{\textwidth}{XX}
    \toprule
    \textbf{Original Synthetic Sub-Theme} & \textbf{GATOS Workflow Method Closest Match(es)} \\
    \midrule
    Setting realistic goals for self-improvement & personal growth; learning and growth; self-reflection \\
    Clear and concise messaging & effective communication; communication effectiveness \\
    Motivates teammates to work towards common goals & effective team dynamics; mutual support and collaboration \\
    Avoiding blame & personal bias management; negative behaviors and their impact \\
    \bottomrule
    \end{tabularx}
    \label{tab:comparison-teammate-feedback-worst}
    \end{table}

\subsubsection{Synthetic Dataset 2: Organizational Culture of Ethics Synthetic Dataset}

With the synthetic dataset of organizational cultures of ethical behavior, there were 61 sub-themes used to generate the original synthetic data. The codebook generated by the GATOS workflow contained 246 codes and 75 themes. An example comparison of the codebook generated by the GATOS workflow with the original sub-themes is shown in Table \ref{tab:comparison-org-ethics-best}. As with the first dataset, for this initial table we have selected the best examples of how the GATOS workflow themes are most clearly aligned with the sub-themes in the original synthetic data. For example, the original sub-theme ``visible consequences for leadership misconduct'' was most closely matched with the GATOS workflow theme ``leadership misconduct and consequences''. Another near-perfect match was for the sub-theme ``regulatory environment constraints'', which was matched with the GATOS workflow theme ``regulatory environment dynamics''.


\begin{table}[htbp]
    \centering
    \fontsize{10}{12}\selectfont
    \caption{Best Comparisons of Codebooks for Organizational Culture of Ethics Synthetic Dataset}
    \begin{tabularx}{\textwidth}{XX}
    \toprule
    \textbf{Original Synthetic Sub-Theme} & \textbf{GATOS Workflow Method Closest Match} \\
    \midrule
    visible consequences for leadership misconduct & leadership misconduct and consequences \\
    addressing microaggressions and harassment & harassment and microaggressions \\
    representation in leadership positions & diversity and representation in leadership \\
    regulatory environment constraints & regulatory environment dynamics \\
    prioritization of ethics in performance evaluations & integration of ethics into performance evaluations \\
    \bottomrule
    \end{tabularx}
    \label{tab:comparison-org-ethics-best}
    \end{table}

Once more, we also present the instances where there were no ideal matches between the GATOS workflow themes and the original sub-themes in Table \ref{tab:comparison-org-ethics-worst}. The specific examples of worst fit between the GATOS workflow-generated themes for this dataset and original sub-themes included ``financial rewards for whistleblowing'', and ``media scrutiny and public opinion''. The nearest matches for the original whistleblower sub-theme alluded to financial rewards whereas the theme generated from the workflow was more general about whistleblowing and retaliation and omitted reference to rewards. Likewise, the media scrutiny and public opinion sub-theme was matched with the themes of (1) public perception and image and (2) media and public relations, which were close but not individually exact matches; however, together the two workflow-generated themes could capture the central idea of the original sub-theme. 


\begin{table}[htbp]
    \centering
    \fontsize{10}{12}\selectfont
    \caption{Worst Comparisons of Codebooks for Organizational Culture of Ethics Synthetic Dataset}
    \begin{tabularx}{\textwidth}{XX}
    \toprule
    \textbf{Original Synthetic Sub-Theme} & \textbf{GATOS Workflow Method Closest Match} \\
    \midrule
    financial rewards for whistleblowing & whistleblowing and retaliation; reporting unethical behavior \\        
    media scrutiny and public opinion & public perception and image; media and public relations \\
    \bottomrule
    \end{tabularx}
    \label{tab:comparison-org-ethics-worst}
    \end{table}

\subsubsection{Synthetic Dataset 3: Perspectives About Returning to Workplace After the Pandemic Synthetic Dataset}

Finally, with the synthetic dataset of returning to the workplace after the pandemic, there were 63 sub-themes used to generate the original synthetic data. The codebook generated by the GATOS workflow contained 314 codes and 110 themes. An example comparison of the codebook generated by the GATOS workflow with the original sub-themes is shown in Table \ref{tab:comparison-return-to-work-best}. As with the other two datasets, there were several examples of near-perfect matches between the original sub-themes and the GATOS workflow-generated themes. For example, the original sub-theme ``resistance to traditional office hours'' was most closely matched with the GATOS workflow theme ``resistance to traditional office hours''. Another near-perfect match was for the sub-theme ``concerns about micromanaging and surveillance'', which was matched with the GATOS workflow theme ``micromanagement and surveillance concerns''. In those instances, even the syntax of the sub-theme was captured in the GATOS workflow-generated theme along with the semantics.   


\begin{table}[htbp]
    \centering
    \fontsize{10}{12}\selectfont
    \caption{Best Comparisons of Codebooks for Returning to Workplace After the Pandemic Synthetic Dataset}
    \begin{tabularx}{\textwidth}{XX}
    \toprule
    \textbf{Original Synthetic Sub-Theme} & \textbf{GATOS Workflow Method} \\
    \midrule
    resistance to traditional office hours & resistance to traditional office hours \\
    concerns about micromanaging and surveillance & micromanagement and surveillance concerns \\
    uncertainty about emergency response plans and protocols & emergency response plan uncertainty \\
    more comfortable and personalized workspace & comfortable workspace customization \\
    nostalgia for office culture and traditions & nostalgia for pre pandemic office culture and traditions \\
    \bottomrule
    \end{tabularx}
    \label{tab:comparison-return-to-work-best}
\end{table}

As with the other two datasets, we also present the instances where there were no ideal matches between the GATOS workflow themes and the original sub-themes in Table \ref{tab:comparison-return-to-work-worst}. Unlike the first two synthetic datasets, however, there were no instances where the GATOS workflow themes did not match the original sub-themes. The worst match was for the sub-theme ``confusion over new policies and procedures'', which was matched with the GATOS workflow theme ``uncertainty about specific post-pandemic office aspects''. Those are close matches, with the difference possibly being the lack of confusion being mentioned in the GATOS workflow theme. The second sub-optimal match was for the sub-theme ``worry about being judged or evaluated'', which was matched with the GATOS workflow themes ``social expectation pressure anxiety''. The aspect possibly missing in the workflow-generated theme is an element of judgment.


\begin{table}[htbp]
    \centering
    \fontsize{10}{12}\selectfont
    \caption{Worst Comparisons of Codebooks for Returning to Workplace After the Pandemic Synthetic Dataset}
    \begin{tabularx}{\textwidth}{XX}
    \toprule
    \textbf{Original Synthetic Sub-Theme} & \textbf{GATOS Workflow Method Closest Match} \\
    \midrule
    confusion over new policies and procedures & uncertainty about specific post pandemic office aspects \\
    worry about being judged or evaluated & social expectation pressure anxiety \\
    \bottomrule
    \end{tabularx}
    \label{tab:comparison-return-to-work-worst}
    \end{table}

\section{Discussion}

Our investigations of the GATOS workflow in this simulation study have shown that the method can generate themes that closely match the sub-themes used to generate the original synthetic data. We borrowed from the practice in quantitative research methods development in which researchers generate synthetic datasets to test the performance of a new method by observing whether the new method captures the underlying data generation process \cite{morris2019using,burton2006design}. In our case, we generated synthetic qualitative data with known themes and sub-themes to test the GATOS workflow. In all three datasets, the GATOS workflow-generated themes were able to capture the essence of the original sub-themes. There were fewer than 5\% of the original sub-themes that did not have a clear match and under 2\% that had no good match. If these results hold in other contexts and datasets from human participants (instead of synthetic datasets), the GATOS workflow could be a valuable tool for researchers to use in qualitative data analysis. 

The plots of code generation rates over time for each dataset (Figures \ref{fig:teammate-feedback-codes-vs-cluster}, \ref{fig:org-culture-codes-vs-cluster}, and \ref{fig:office-return-codes-vs-cluster}) show that the GATOS workflow was able to generate fewer new codes as it encountered more clusters. For us, this was only a recently observed phenomena with newer generative text models because prior models would not consistently generate internally coherent reasoning traces to decide whether or not to create a new code. Anecdotally, in our experience with prior versions of generative text models trained on fewer data, the models would display faulty reasoning 10-20\% of the time. That error rate was unacceptable for applications in research. However, the newer generations of models (e.g., Mistral-small-22b) have improved to the point where they can consistently generate plausible and consistent reasoning. This level of reliability is crucial for any method that is intended to be used in research \cite{clonts1992concept,golafshani2003understanding}.

\subsection{Comparison with Other Studies Using Generative Text Models for Qualitative Data Analysis}
The idea of using generative text models to facilitate qualitative data analysis is not new. Yet, the GATOS workflow is unique in several ways. First, the GATOS workflow uses open-source generative text models, which makes the method more accessible to researchers. Many prior studies have used proprietary generative text models (e.g., ChatGPT), which can be expensive and difficult to access. Using those proprietary models also raises important ethical concerns regarding data privacy and security \cite{davison2024ethics}. Many of these concerns arise from sending data to third-party model providers where the research team no longer has control over the data. Instead, using local, permissively licensed (e.g., Apache-2.0 license) LLMs can help to mitigate those concerns by avoiding such data transmission. Using open-source models also enables better control over which model is used. To wit, when using proprietary models, researchers are often limited to the models that are available from the provider, which can be updated or changed without a researcher realizing the change. This is detrimental to the reproducibility of the research because different models yield different results, as evidenced by the constant leaderboard changes and performance metric saturation \cite{ott2022mapping}. When an upgrade happens to a model used in a research study, this might raise issues similar to inter-rater reliability with human research teams \cite{belotto2018data}. Although these language models are probabilistic and might have seemingly inherent limitations in their ability to provide consistency, we addressed this issue by setting the model temperature to 0. This setting forces a more deterministic output for a model when given the same input.

Second, our approach develops an entire workflow for qualitative data analysis, from initial summarization to theme creation. Many prior studies have only used generative text models for one part of the qualitative data analysis process, such as coding or summarization. For example, Prescott et al. \cite{prescott2024comparing} simply sent their data to ChatGPT and Bard. In a different study, Perkins and Roe \cite{perkins2024use} used a human-in-the-loop approach to work with ChatGPT through multiple rounds of prompting to assist their inductive qualitative analysis. Qiao et al. \cite{qiao2024generative} took yet a different approach and fine-tuned ChatGPT before then sending their data to their fine-tuned ChatGPT to perform various steps of thematic analysis. The GATOS workflow is different from these approaches in several regards. First, the workflow uses a process more common in qualitative codebook generation with several steps that align with Braun and Clarke's steps for thematic analysis \cite{braun2006using}. Second, the workflow uses a generative text model at multiple steps (summarization, codebook generation, and theme creation) along with an embedding model to add some structure to otherwise unstructured data. This structuring is important for the workflow to be able to generate themes that are coherent and meaningful and capture recurring patterns in the data. Third, referring to the point in the preceding paragraph, the GATOS workflow uses open-source generative text models rather than ChatGPT (or other proprietary models), which makes the method more accessible to researchers and alleviates some concerns about data security. While some researchers advocate for using ChatGPT in social science research \cite{salah2023may}, we would caution against that approach because of the aforementioned concerns about data privacy and model transparency, among others.

Another contribution of our work compared to other recent studies that used ChatGPT for qualitative data analysis is our use of synthetic datasets to demonstrate the utility of qualitative text analysis method. Generating and then testing a qualitative method on simulated data is an approach that allows researchers to test the performance of the method in a controlled setting. Simulation studies are common in quantitative research methods development \cite{morris2019using}, but they are much less common in qualitative research methods development. This might in part be due to ontological and epistemological assumptions associated with qualitative paradigms, which often emphasize the importance of context and the situatedness of knowledge \cite{guba1994competing,lincoln2011paradigmatic}. As others have noted, ontology and epistemology precede methodology \cite{grix2002introducing}. We do not mean to suggest the GATOS workflow will be most consistent with some researchers' worldviews; yet, for those interested in computer-assisted approaches or researchers grappling with large volumes of qualitative data, the GATOS workflow could be a valuable tool.

Along with mimicking the simulation study paradigm from quantitative research traditions, using synthetic data also enables another advantage: working at a larger scale. Many of the extant studies using generative text models for qualitative data analysis worked on much smaller scales. Prescott et al. used 40 SMS text messages. Qiao et al. used nine interviews. Perkins and Roe used 28 policy documents. These are all relatively small datasets. The GATOS workflow, on the other hand, was able to work with datasets of 854, 823, and 1,110 observations. This is a much larger scale than many other studies using generative text models for qualitative data analysis. In our view, the ability to work at this scale is one of the main reasons to use computer-assisted analysis techniques. When conducting organizational research or other kinds of social science research, working at that scale can be resource-intensive for a team of human researchers. The GATOS workflow can help to alleviate some of those resource constraints and enable researchers to work with larger datasets more efficiently. In turn, working at that scale can allow researchers to answer more systematic questions about organizational phenomena, thereby trying to find a middle ground between the richness of qualitative traditions and the statistical power of quantitative traditions.

\subsection{Future Research}
There are several natural extensions to the work we presented here introducing the GATOS workflow. Some of the most immediate future research directions include: testing other datasets that have already been analyzed in a more traditional qualitative coding process; testing other language models, embedding models, and cluster analysis methods for applying the codebook; testing the scalability of the method to larger datasets; and testing the method on other data with more complex structures, e.g., long interview transcripts. The first extension - trying to use the GATOS workflow to replicate human analysis on live datasets - would be a good test of the method's generalizability and the extent to which it can mimic what human researchers had previously done. Indeed, this is a study currently underway by the authors on three datasets collected from human participants in educational research settings. 

The second extension - testing other language models, embedding models, and cluster analysis methods - would be a good test of the method's flexibility and robustness, akin to sensitivity analyses. In typical quantitative research sensitivity analyses, researchers test the robustness of their results to different assumptions or methods \cite{saltelli2000sensitivity,iooss2015review}. When there are complex systems and models being used for analysis, there are often myriad assumptions and parameters that could be affecting the output. Systematically testing the method with different language models, embedding models, and cluster analysis methods would help to ensure that the method is robust to those choices. Such analysis could also function as an optimization search for the best combination of models and methods to use in the GATOS workflow.

Finally, testing the model on larger quantities of data and more complex data structures would help identify the boundaries of where the GATOS workflow does and does not work well. For example, the simulated data tested for this study were relatively short responses to a single question. Such data tend to represent a small number of topics. In contrast, semi-structured interviews can explore a much larger topic space, depending on the phenomena investigated and questions asked. In our initial testing, we have explored segmenting interviews into smaller units of analysis, but we have not yet tested the GATOS workflow on full interview transcripts. Doing so would be a fruitful area of future research.

\subsection{Limitations}
The present study contains several limitations. First, the work we presented here utilized synthetic data for validation. There is a tradeoff in that approach. On the one hand, this allows for control over the ``ground truth'' that the workflow should recover. We presented this as a significant contribution earlier in this Discussion section. On the other hand, the synthetic data may not fully capture the complexity and nuances of human-generated qualitative data. Additionally, there could be something about the way the data were generated that keeps them in distribution for subsequent language models to summarize, embed, and analyze, which is unlike live data. To mitigate part of this issue, we used different language models for the data synthesis and codebook generation tasks. Specifying personas, contexts, and writing styles in the data generation criteria was also designed to help this issue. We also read a subset of each synthetic dataset to ensure that the data were coherent and plausible. Second, generative text models have been noted to have some biases when completing select tasks \cite{liang2021towards}. This is especially noted in generative text-to-image models \cite{bird2023typology}. While we did not observe any biases in the generative text models we used, it is possible that biases could emerge in other contexts. 

Regarding the workflow itself, there are other notable limitations. First, the workflow still generates nearly redundant codes sometimes. The move from codes to themes helps to reduce those redundancies, but they occasionally persisted nonetheless. The extent to which this is a concern depends on the use case. Second, there are open questions about the appropriate level of abstraction for the code generation. This is an area where we believe a human-in-the-loop approach can help the workflow and refine the final codebook. Third, there remain open questions about the scalability of the method. The steps in the GATOS workflow can be computationally intensive when working with tens of thousands of observations. For example, imagine there are 10,000 comments that one has collected in a large-scale study of corporate culture. The GATOS workflow would need to first summarize each of those comments. If we assume that each comment leads to five summary points, then we now have 50,000 summary points to cluster. Assuming that each cluster has 10 summary points, that still leads to 5,000 clusters for the model to analyze when determining whether to create a new code. This is a computationally intensive process, and it is not clear how well the method would scale to even larger datasets. One way to address this issue would be to use iterative rounds of clustering the summary points to reduce instances where there are clear repetitions of summary points because people said the same thing. This would reduce the number of clusters the model would need to analyze.

\section{Conclusion}

In this paper, we have presented a method to use open-source NLP tools and generative text models to mimic thematic analysis of qualitative data. We call this method the Generative AI-enabled Theme Organization and Structuring (GATOS) workflow. To test the ability of the GATOS workflow to accurately capture themes and sub-themes in qualitative data, we generated three synthetic datasets to mimic scenarios one might investigate in organizational research. We then applied the GATOS workflow to these simulated datasets and compared the themes generated by the workflow to the sub-themes used to generate the original synthetic data. We found that the GATOS workflow was able to capture the themes and sub-themes in the synthetic data with a high degree of accuracy. The workflow was able to generate themes that closely matched the original sub-themes in the synthetic data, with only a few instances where the match was not ideal. These findings have implications for organizational research and other fields that rely on qualitative data analysis. The GATOS workflow can support scalable qualitative data analysis and provide a new tool for researchers to analyze large volumes of qualitative data. Future research should explore the scalability of the method to larger datasets and test the method on other data with more complex structures. We believe the GATOS workflow has the potential to transform qualitative data analysis and support researchers in generating insights from qualitative data in a more efficient and effective way.

\section{Acknowledgements} 
This work was supported by the National Science Foundation under Grant No. 2107008 and the Virginia Tech Academy of Data Science Discovery Fund. The authors would like to thank the reviewers and colleagues whose comments and suggestions on this idea have been invaluable.

\section{Appendix}
\label{sec:appendix}

\subsection{Prompts}
This section contains the prompts used for the steps in the GATOS workflow. 
\subsection{Initial Summarization Prompt}
\label{subsec:initial-summarization-prompt}

\begin{tcolorbox}[colback=gray!10, colframe=gray!80, title=Information Extraction Instructions, boxrule=0.5mm, left=1mm, right=1mm]
    \fontsize{10}{12}\selectfont
    \lstset{
        basicstyle=\normalfont\small,
        breaklines=true,
        columns=fullflexible,
        frame=none,
        backgroundcolor=\color{gray!10},
        xleftmargin=0mm,
        xrightmargin=0mm,
        numbers=none,
        showspaces=false,
        breakindent=0pt
    }
    \begin{lstlisting}[breaklines=true]
You are an expert text analyst reading {data_type}s collected in {data_collection_context}. I am going to send you one of these {data_type}s. I need you to use your expertise to analyze the provided text in the <text> tag below and summarize it in an enumerated list. You should do this analysis by providing several short descriptive phrases that summarize each idea discussed in the {data_type} that answered the prompt. When you suggest multiple items, separate each one in your response with a new line. You MUST remove anyone's names and *use gender neutral pronouns* for deidentification purposes. Start your response with ``My summary:". Here is an example of input and desired output from a different context when there are only two topics, but remember that you can suggest as many topics as you think are necessary for the text you summarize.
Example input: ``Jared did a great job responding quickly to emails and turning in good work.''
Example output: ``My summary: 
1. Responded quickly to emails 
2. Turned in good work''.
Notice how the main ideas are summarized and there are no names or pronouns included here. Also, notice how the response did not make up information that was not in the input. You must NEVER make up information that is not in the input text you receive because there is a severe penalty for that. If the text you receive is very short and says ``nothing'', do not make up new things.
Here is the text for you to summarize: <text>{text}</text> 
Begin your analysis now.
    \end{lstlisting}
\end{tcolorbox}

\subsection{Initial Codebook Creation Prompt}
\label{subsec:cb-create-prompt}

\begin{tcolorbox}[colback=gray!10, colframe=gray!80, title=Initial Codebook Creation, boxrule=0.5mm, left=1mm, right=1mm]
    \fontsize{10}{12}\selectfont
    \lstset{
        basicstyle=\normalfont\small,
        breaklines=true,
        columns=fullflexible,
        frame=none,
        backgroundcolor=\color{gray!10},
        xleftmargin=0mm,
        xrightmargin=0mm,
        numbers=none,
        showspaces=false,
        breakindent=0pt
    }
    \begin{lstlisting}[basicstyle=\small, breaklines=true]
Act as if you are the world's best qualitative data analysis. You specialize in applying codes to analyze qualitative data. I need your help. Your important task is to generate {k_to_start} hypothetical codes that one might encounter when analyzing {data_type}s from {data_collection_context}. You should format your response by filling in the template I give you at the end of these instructions, which is an enumerated list of {k_to_start} codes. The list should contain {k_to_start} short phrases with regular spacing between words written in plain English without examples. After the final code, you should stop writing so that it is easy for your response to be parsed for downstream tasks. Begin your list now using the following template:\n{code_template}
    \end{lstlisting}
\end{tcolorbox}

where code template is a numbered string of k\_to\_start codes. If k\_to\_start is 5, then the code template would be
1. Code 1
2. Code 2
3. Code 3
4. Code 4
5. Code 5

\subsection{Inductive Codebook Generation Prompt}
\label{subsec:cb-ind-gen-prompt}
The inductive codebook generation prompt contains multiple parts. The first part introduces the persona for the model to adopt for this task. Prior research suggests persona assignment can be a way to improve generative text model performance \cite{olea2024evaluating,hu2024quantifying}. The next part of this initial portion of the prompt provides the model with the background information about the task.  
\begin{tcolorbox}[colback=gray!10, colframe=gray!80, title=Inductive Codebook Generation, boxrule=0.5mm, left=1mm, right=1mm]
    \fontsize{10}{12}\selectfont
    \lstset{
        basicstyle=\normalfont\small,
        breaklines=true,
        columns=fullflexible,
        frame=none,
        backgroundcolor=\color{gray!10},
        xleftmargin=0mm,
        xrightmargin=0mm,
        numbers=none,
        showspaces=false,
        breakindent=0pt
    }
    \begin{lstlisting}[basicstyle=\small, breaklines=true]
        
Act as if you are the world's best qualitative data analyst with expertise in generating qualitative codebooks for thematic analysis. You specialize in creating parsimonious codebooks with non-overlapping and non-redundant codes. A codebook in this setting is a collection of labels and definitions for those labels that can be used to describe pieces of data in a qualitative research study. I need your help to create a qualitative codebook to analyze {data_type}s from {data_collection_context}. To aid you in this process, I am going to send you instructions in the <instructions> XML tag. Use the instructions to analyze the data in the <data_to_analyze> tag. You must follow these instructions using your expertise and data to analyze in the <data_to_analyze> XML tag. I will provide you the instructions first and then the data to analyze afterward. Be aware that your instructions contain task instructions, evaluation criteria, and formatting instructions, each in their respective XML tags.
\end{lstlisting}
\end{tcolorbox}

The next portion of the prompt introduces the task instructions for the model to follow.
\begin{tcolorbox}[colback=gray!10, colframe=gray!80, title=Instructions for Inductive Codebook Generation, boxrule=0.5mm, left=1mm, right=1mm]
    \fontsize{10}{12}\selectfont
    \lstset{
        basicstyle=\normalfont\small,
        breaklines=true,
        columns=fullflexible,
        frame=none,
        backgroundcolor=\color{gray!10},
        xleftmargin=0mm,
        xrightmargin=0mm,
        numbers=none,
        showspaces=false,
        breakindent=0pt
    }
    \begin{lstlisting}[basicstyle=\small, breaklines=true]
<instructions>
<task_instructions>
We are trying to determine whether or not an exsiting codebook is sufficient for analyzing one {data_type} that you have been given in the <text_to_analyze> tag. Your important task is to analyze one summary of {data_type}s collected in the context of {data_collection_context} and determine if the theme discussed in the {data_type} summary is already covered by the codes in an existing codebook that will be given to you in the <existing_codebook> tag or if instead the codebook needs one or more new code to cover the theme in the text to analyze. You should complete your task by following these steps:

\end{lstlisting}
\end{tcolorbox}

\begin{tcolorbox}[colback=gray!10, colframe=gray!80, title=Steps 1-4 for Inductive Codebook Generation, boxrule=0.5mm, left=1mm, right=1mm]
    \fontsize{10}{12}\selectfont
    \lstset{
        basicstyle=\normalfont\small,
        breaklines=true,
        columns=fullflexible,
        frame=none,
        backgroundcolor=\color{gray!10},
        xleftmargin=0mm,
        xrightmargin=0mm,
        numbers=none,
        showspaces=false,
        breakindent=0pt
    }
    \begin{lstlisting}[basicstyle=\small, breaklines=true]
Step 1: Read existing codebook. 
Examine the existing codebook given to you in the <existing_codebook> tag. Describe what these codes are discussing.
Step 2: Read the summary of the {data_type}s. 
Read the new summary of the {data_type} given to you in the <text_to_analyze> tag and identify the main theme discussed in the summary.
Step 3: Try to use existing codebook.
Attempt to describe the main theme of the {data_type} using one or more of the existing codes in the existing codebook. Think at a high level of abstraction and consider if any new themes could be subcategories of existing codes. If you determine that there is no need to create a new code, say "No new codes needed".
Step 4: Create new code if needed.
If in step 3 you discover that you are unable to use the current codes to describe the main theme in the summary of the {data_type} that you are analyzing, determine whether the existing codebook needs new labels to describe the summary in the <text_to_analyze> tag. You should complete this determination by reasoning step-by-step. If you determine that a new code is necessary, explicitly justify why existing codes or combinations thereof are insufficient. Finally, generate a new code (or codes, if multiple ones are absolutely necessary) that captures the main concepts or themes discussed in the {data_type}s that you review. Remember, you specialize in creating parsimonious codebooks and avoid creating redundant codes. Your goal is to use the least number of new codes possible while still accurately representing the data.
There is a VERY significant penalty for creating redundant or unnecessary codes, so you should only create a new code if you are **absolutely** certain the existing ones are insufficient, even when combined or broadened. If you decide to generate a new code, please provide:
- The code (a short phrase).
- A brief definition of what the label represents.
\end{lstlisting}
\end{tcolorbox}

\begin{tcolorbox}[colback=gray!10, colframe=gray!80, title=Steps 5-6 for Inductive Codebook Generation, boxrule=0.5mm, left=1mm, right=1mm]
    \fontsize{10}{12}\selectfont
    \lstset{
        basicstyle=\normalfont\small,
        breaklines=true,
        columns=fullflexible,
        frame=none,
        backgroundcolor=\color{gray!10},
        xleftmargin=0mm,
        xrightmargin=0mm,
        numbers=none,
        showspaces=false,
        breakindent=0pt
    }
    \begin{lstlisting}[basicstyle=\small, breaklines=true]
Step 5: Evaluate your suggestion.
To guide your work, you must consider the following three evaluation criteria. These three evaluation criteria will be used by other famous expert qualitative data analysts to evaluate the quality of your work. In the reflection step, you must check whether you have satisfied each of these three criteria:
<evaluation_criteria>
Evaluation Criteria 1. Parsimony: Have you made every effort to use existing codes or combinations of existing codes before proposing a new one?
Evaluation Criteria 2. Abstraction Level: Is any proposed new code at an appropriate level of abstraction, consistent with existing codes?
Evaluation Criteria 3. Non-Redundancy: Have you avoided creating codes that significantly overlap with existing ones?

To help illustrate what I mean by non-redundancy, here is an example of redundant codes and an explanation of their redundancy:
{redundancy_example}

Use the evaluation criteria and these task instructions to help you in your step-by-step reasoning for each of the preparation, analysis, and reflection steps given to you in these instructions. 
It is CRUCIAL TO REMEMBER that if you do not think a new code should be created, you must say "No new codes needed". 
</evaluation_criteria>
Step 6: Final recommendation.
Present your final logical recommendation on a new line about any codes to create or whether none are needed on a new line.
</task_instructions>

\end{lstlisting}
\end{tcolorbox}

\begin{tcolorbox}[colback=gray!10, colframe=gray!80, title=Formatting Instructions for Inductive Codebook Generation, boxrule=0.5mm, left=1mm, right=1mm]
    \fontsize{10}{12}\selectfont
    \lstset{
        basicstyle=\normalfont\small,
        breaklines=true,
        columns=fullflexible,
        frame=none,
        backgroundcolor=\color{gray!10},
        xleftmargin=0mm,
        xrightmargin=0mm,
        numbers=none,
        showspaces=false,
        breakindent=0pt
    }
    \begin{lstlisting}[basicstyle=\small, breaklines=true]
<formatting_instructions>
I will give you a template to use for your response. The main parts of the template are the following. First, your response should start with "My expert analysis:". Then, on a new line, you should write your logical step-by-step reasoning about the existing codes and the {data_type}s. This will include the two orientation steps, the two analysis steps, the reflection step, and the recommendation step. Your anlayis notes should be succinct and formatted in a numbered list rather than long prose. This means that each step in your step-by-step reasoning should get its own line as if it were a premise in a proof. These notes should be logical, adhere perfectly to your task instructions, be concise, and be in a numbered list. Then, on another new line, you should state "My logical recommendation:" followed by your recommendation on yet another new line. Your recommendations can either be "No new codes needed" if no new codes are needed or the actual codes you suggest adding to the codebook. If you do think one or more new codes should be created, your response should start 'Code: ' followed by your code, then on a new line 'Definition: ' followed by your definition for that code.
For example:
Code: <code 1>
Definition: <definition 1>
</formatting_instructions>
This concludes your task and formatting instructions.
</instructions>
        
\end{lstlisting}
\end{tcolorbox}

\begin{tcolorbox}[colback=gray!10, colframe=gray!80, title=Presenting Data and Template for Inductive Codebook Generation, boxrule=0.5mm, left=1mm, right=1mm]
    \fontsize{10}{12}\selectfont
    \lstset{
        basicstyle=\normalfont\small,
        breaklines=true,
        columns=fullflexible,
        frame=none,
        backgroundcolor=\color{gray!10},
        xleftmargin=0mm,
        xrightmargin=0mm,
        numbers=none,
        showspaces=false,
        breakindent=0pt
    }
    \begin{lstlisting}[basicstyle=\small, breaklines=true]
Now that you have studied your instructions, here are the data for you to analyze.

<data_to_analyze>
<existing codebook>
{codes}
</existing codebook>
And here is a summary of one {data_type} for you to analyze.
<text_to_analyze>
{text}
</text_to_analyze>
</data_to_analyze>

Now that you have meticulously studied the data to analyze using your task instructions, formatting instructions, and evaluation criteria, take a moment to gather your expert thoughts and observations. When you are ready, begin your flawless and logical step-by-step analysis using the instructions and evaluation criteria outlined above. Be sure to display your expertise in creating parsimonious codebooks and minimizing redundancy and use the full analysis template, provided below. Be sure to use spaces in any codes you write rather than concatenating words together (e.g., say "example code" rather than "examplecode"). Here is the template to use for your analysis. Begin your expert analysis when you are ready.
FULL ANALYSIS TEMPLATE:
My expert analysis:
Step 1 (codebook examination)
[your step 1 notes describing the existing code go here]
Step 2 (current data examination)
[your step 2 notes go here to identify the main theme in the {data_type}]
Step 3 (analysis part 1)
[your step 3 notes to describe main theme in the {data_type}s with existing codes here]
Step 4 (analysis part 2)
[your step 4 notes considering whether  to create new code here, favoring parsimony and avoiding unnecessary code creation]
Step 5 (reflection on planned suggestions)
[your evaluation reflection notes here reviewing the evaluation criteria]
My logical recommendation:
[logical recommendation based on expert step-by-step reasoning about whether or not to create zero, one, or more than one new codes. These notes will reflect your reputation for only creating essential codes]
\end{lstlisting}
\end{tcolorbox}

\subsection{Theme Identification Prompt}
\label{subsec:cb-theme-identification-prompt}

\begin{tcolorbox}[colback=gray!10, colframe=gray!80, title=Theme Identification Instructions, boxrule=0.5mm, left=1mm, right=1mm]
    \fontsize{10}{12}\selectfont
    \lstset{
        basicstyle=\normalfont\small,
        breaklines=true,
        columns=fullflexible,
        frame=none,
        backgroundcolor=\color{gray!10},
        xleftmargin=0mm,
        xrightmargin=0mm,
        numbers=none,
        showspaces=false,
        breakindent=0pt
    }
    \begin{lstlisting}[basicstyle=\small, breaklines=true]
        You are an expert qualitative researcher specializing in thematic analysis. Your task is to analyze a list of codes that will be given to you below in the <codes> tag and identify potential themes following the guidance of Braun and Clarke. The goal is to identify themes that help to answer the research question ``{research_question}''. Please follow these steps outlined in the <instructions> tag carefully.
        \end{lstlisting}
\end{tcolorbox}

The specific instructions for the theme identification task are provided in the following box.
\begin{tcolorbox}[colback=gray!10, colframe=gray!80, title=Instructions for Theme Identification, boxrule=0.5mm, left=1mm, right=1mm]
    \fontsize{10}{12}\selectfont
    \lstset{
        basicstyle=\normalfont\small,
        breaklines=true,
        columns=fullflexible,
        frame=none,
        backgroundcolor=\color{gray!10},
        xleftmargin=0mm,
        xrightmargin=0mm,
        numbers=none,
        showspaces=false,
        breakindent=0pt
    }
    \begin{lstlisting}[basicstyle=\small, breaklines=true]   
        <instructions>
        Step 1. Review the list of codes provided below in the <codes> tag below. These codes are being used to analyze {data_type}s from {data_collection_context}.
        Step 2. Look for patterns and shared meanings among the codes. Consider how different codes might be combined based on underlying concepts or features of the data.
        Step 3. Identify overarching narratives that might represent broader themes or sub-themes.
        Step 4. Remember that themes don't simply "emerge" from the data. Actively construe relationships among the codes and examine how these relationships inform potential themes.
        Step 5. Consider the importance and salience of potential themes. Remember, the number of codes supporting a theme is less important than whether the pattern communicates something meaningful that helps answer the research question(s). On that note, remember that the research question for this research is {research_question}.
        Step 6. Aim for themes that are distinctive yet coherent with the overall analysis. Themes may even be contradictory to each other.
        Step 7. Be willing to let go of codes or potential themes that don't fit the overall analysis. Consider creating a "miscellaneous" category for codes that don't fit elsewhere.
        Step 8. Strive for a balance in the number of themes - not so many that the analysis becomes unwieldy, but enough to fully explore the depth and breadth of the data.
        Step 9. For each theme, prepare a structured description including the theme name, its underlying concept, associated codes, and how these codes relate to each other and the overall theme.
        Step 10. Reflect on your analysis considering: themes that seem too broad or narrow, contradictions or unexpected patterns, need for subthemes, and codes that don't fit well into the current themes.
        Step 11. Organize your analysis into a structured format with initial observations, an array of suggested themes (each as an object with name, concept, codes, and relationship), and your reflection.
        </instructions>
    \end{lstlisting}
\end{tcolorbox}

The prompt also includes a section for the model to provide the list of codes to analyze.
\begin{tcolorbox}[colback=gray!10, colframe=gray!80, title=Presenting Data and Template for Theme Identification, boxrule=0.5mm, left=1mm, right=1mm]
    \fontsize{10}{12}\selectfont
    \lstset{
        basicstyle=\normalfont\small,
        breaklines=true,
        columns=fullflexible,
        frame=none,
        backgroundcolor=\color{gray!10},
        xleftmargin=0mm,
        xrightmargin=0mm,
        numbers=none,
        showspaces=false,
        breakindent=0pt
    }
    \begin{lstlisting}[basicstyle=\small, breaklines=true]        
        Now that you have studied your instructions carefully, here is the list of codes to analyze to identify themes related to the research question "{research_question}":
        <codes>
        {labels}
        </codes>
        
        Proceed with your expert analysis, explaining your reasoning at each step. Present your analysis in JSON format with the following structure:
        
        {{
          "initial_observations": [
            "observation1"
          ],
          "suggested_themes": [
            {{
              "theme_name": "Theme 1",
              "concept": "Brief description of the underlying concept or narrative",
              "codes": [
                "Code 1"
              ],
              "relationship": "Brief explanation of how these codes relate to each other and the overall theme"
            }}
          ],
          "reflection": {{
            "broad_or_narrow_themes": "Discussion of any themes that seem too broad or too narrow",
            "contradictions_or_unexpected_patterns": "Description of any contradictions or unexpected patterns",
            "potential_subthemes": "Discussion of any need for subthemes within the main themes",
            "unclassified_codes": "List of any codes that were not included in the proposed themes"
          }}
        }}
        
        Use this JSON structure I have given you as a template. Expand on the template by adding as many observations, themes, and codes as necessary based on your analysis. Ensure that your response remains a valid JSON object. Do not include any text outside of this JSON structure.
        
        Now that you have thoroughly read your task instructions, formatting instructions, and the codes to analyze, take a moment to gather your expert thoughts. Begin your analysis when you are ready.
    \end{lstlisting}
\end{tcolorbox}

\bibliographystyle{IEEEtran}


\end{document}